\documentclass{article}

 \usepackage[preprint]{neurips_2025}

\usepackage[utf8]{inputenc} % allow utf-8 input
\usepackage[T1]{fontenc}    % use 8-bit T1 fonts
\usepackage{hyperref}       % hyperlinks
\usepackage{url}            % simple URL typesetting
\usepackage{booktabs}       % professional-quality tables
\usepackage{amsfonts}       % blackboard math symbols
\usepackage{nicefrac}       % compact symbols for 1/2, etc.
\usepackage{microtype}      % microtypography
\usepackage{subcaption}
\usepackage{enumitem}
\usepackage{graphicx}
\usepackage{amsmath}
\usepackage{multirow}
\usepackage{wrapfig}
\title{\textit{XStreamVGGT}: Extremely Memory-Efficient Streaming Vision Geometry Grounded Transformer with KV Cache Compression}

\author{
  Zunhai Su$^{1*}$\enskip\enskip
  Weihao Ye$^{2}$\thanks{Equal Contribution. }\enskip\enskip
  Hansen Feng$^{1}$\enskip\enskip
  Keyu Fan$^{1}$\enskip\enskip\\
  \vspace{2mm}
  \textbf{Jing Zhang$^{3}$\enskip\enskip
  Dahai Yu$^{4}$\enskip\enskip
  Zhengwu Liu$^{5}$\enskip\enskip
  Ngai Wong$^{5}$}\\
  $^{1}$Shenzhen International Graduate School, Tsinghua University\\
  $^{2}$Institute of Artificial Intelligence, Xiamen University\\
  $^{3}$China Star Optoelectronics Technology\\
  $^{4}$TCL Corporate Research (HK) Co., Ltd.\\
  $^{5}$Department of Electrical and Electronic Engineering, The University of Hong Kong\\
}

\begin{document}

\maketitle

\begin{abstract}
Learning-based 3D visual geometry models have significantly advanced with the advent of large-scale transformers.
Among these, StreamVGGT leverages frame-wise causal attention to deliver robust and efficient streaming 3D reconstruction.
However, it suffers from unbounded growth in the Key-Value (KV) cache due to the massive influx of vision tokens from multi-image and long-video inputs, leading to increased memory consumption and inference latency as input frames accumulate. 
This ultimately limits its scalability for long-horizon applications.
To address this gap, we propose XStreamVGGT, a tuning-free approach that seamlessly integrates pruning and quantization to systematically compress the KV cache, enabling extremely memory-efficient streaming inference.
Specifically, redundant KVs generated from multi-frame inputs are initially pruned to conform to a fixed KV memory budget using an efficient token-importance identification mechanism that maintains full compatibility with high-performance attention kernels (e.g., FlashAttention).
Additionally, leveraging the inherent distribution patterns of KV tensors, we apply dimension-adaptive KV quantization within the pruning pipeline to further minimize memory overhead while preserving numerical accuracy.
Extensive evaluations show that XStreamVGGT achieves mostly negligible performance degradation while substantially reducing memory usage by 4.42$\times$ and accelerating inference by 5.48$\times$, enabling practical and scalable streaming 3D applications.
The code is available at \url{https://github.com/ywh187/XStreamVGGT/}.
\end{abstract}

\section{Introduction}
% StreamVGGT and its problems
Recovering 3D geometric structures from image sequences has long been a fundamental challenge in 3D computer vision~\cite{samavati2023deep}. 
This task is crucial for numerous real-world applications, such as robotics, augmented reality, and autonomous driving \cite{robinson2023robotic,dargan2023augmented,chen2022milestones}.
For years, 3D vision has been dominated by classical methods, primarily structure-from-motion (SfM) \cite{agarwal2011building,frahm2010building,wu2013towards} and multi-view stereo (MVS) \cite{gu2020cascade,furukawa2009accurate}.
While these methods demonstrate strong performance in high-fidelity geometric optimization, they rely on fragmented multi-stage pipelines that incur substantial processing delays. 
Such pipelines are also susceptible to cascading errors, potentially compromising the accuracy and integrity of the final reconstruction.
In recent years, learning-based feed-forward models have revolutionized this field. 
% \begin{figure}[t]
\begin{wrapfigure}{r}{0.6\textwidth}
    \centering
    \vspace{-2mm}
    \includegraphics[width=0.6\columnwidth]{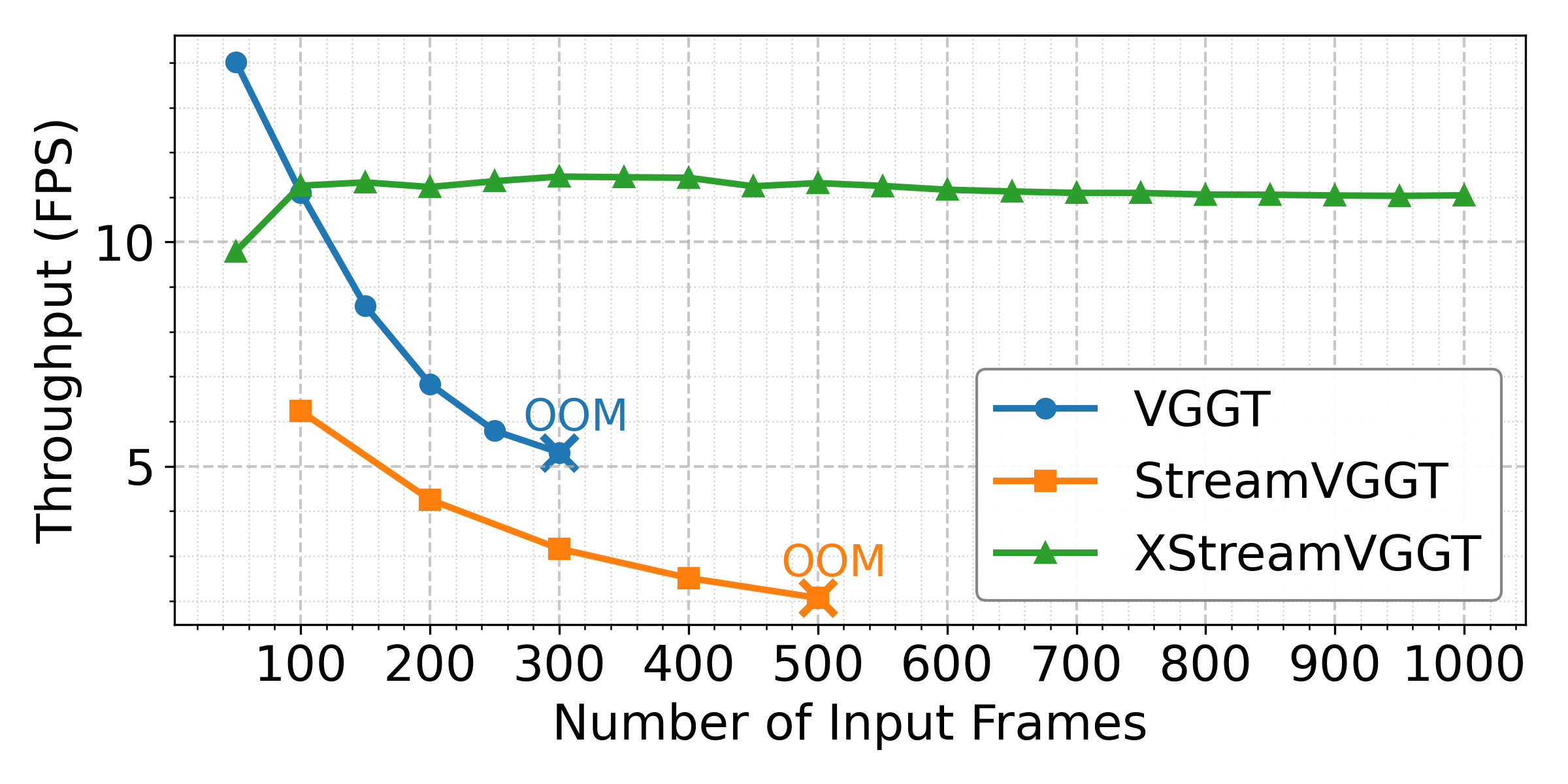}
    \caption{Efficiency analysis on a single 80GB A100 GPU. As the number of input frames increases, StreamVGGT and VGGT exhibit significant FPS degradation and rapidly encounter out-of-memory (OOM) errors. 
    In contrast, XStreamVGGT consistently delivers significantly higher frames per second (FPS) without encountering OOM issues.}
    \vspace{-2mm}
    \label{fig:fps}
\end{wrapfigure}
% \end{figure}
Models such as DUSt3R, CUT3R, and VGGT \cite{wang2025vggt, wang2024dust3r, wang2025continuous} have shifted the paradigm from conventional approaches to end-to-end deep learning frameworks, showcasing not only remarkable performance but also impressive generalization across diverse datasets.
As a significant milestone in this evolution, the Visual Geometry-Grounded Transformer (VGGT) consolidates multiple 3D vision tasks within a unified framework, consistently surpassing task-specific methods across a broad spectrum of applications, including dense depth estimation, point map regression, and camera pose prediction~\cite{wang2025vggt}.
To support robust streaming applications, StreamVGGT \cite{zhuo2025streaming} replaces the global attention mechanism in Alternative-Attention of VGGT with frame-wise causal attention, following a design philosophy similar to autoregressive large language models (LLMs) \cite{team2025longcat1,team2025longcat2,team2026longcat}. 
This change shifts the model from an offline to an online streaming paradigm, substantially improving practical usability.

At the core of StreamVGGT is its reliance on the Key-Value (KV) cache of previous input frames, which functions as an explicit and persistent memory mechanism. 
However, as streaming inference progresses, the model is exposed to a significant influx of vision tokens from multi-image and long-video inputs. 
This influx drives the KV cache to expand linearly with the number of input frames, eventually resulting in unbounded growth \cite{liu2024kivi,hooper2024kvquant}. 
As illustrated in Figure \ref{fig:fps}, memory consumption and inference latency escalate rapidly as input frames accumulate in StreamVGGT. 
This issue significantly restricts the system's scalability for long-horizon applications, creating a critical bottleneck for real-world deployment.

% in this work
To address this challenge, we propose XStreamVGGT, a tuning-free approach that seamlessly integrates pruning and quantization to systematically compress the KV cache, enabling highly memory-efficient streaming inference. 
Specifically, XStreamVGGT first eliminates redundant KV cache from multi-frame inputs through an efficient token importance identification mechanism, pruning the cache to a bounded budget while preserving the first-frame KVs as geometric references~\cite{wang2025vggt}. 
Furthermore, our analysis reveals pronounced channel-wise outliers in the Key tensors, while the Value tensors exhibit much weaker outlier behavior in StreamVGGT. 
Leveraging the inherent distribution patterns of KV tensors, we develop a dimension-adaptive KV quantization scheme that incorporates per-channel Key and per-token Value quantization. 
This quantization scheme is seamlessly integrated into the pruning pipeline, further minimizing memory overhead while maintaining numerical accuracy.
% contribuations
Our key contributions can be summarized as follows:
\begin{itemize}

    \item We introduce XStreamVGGT, the first method to seamlessly integrate pruning and quantization for systematically compressing the KV cache in StreamVGGT. XStreamVGGT effectively addresses the issue of unbounded growth in KV memory, enabling extremely memory-efficient streaming inference.

    \item We conducted a comprehensive analysis of the KV distribution in StreamVGGT, unveiling, for the first time, the distinctive distribution patterns of Key and Value tensors in 3D reconstruction transformer models. 
    By leveraging dimension-adaptive quantization, we effectively mitigate its impact on quantization accuracy.
    
    \item Extensive evaluations on 3D reconstruction, camera pose estimation and depth estimation demonstrate that XStreamVGGT achieves mostly negligible performance degradation, while reducing memory usage by 4.42$\times$ and accelerating inference by 5.48$\times$, offering a powerful and scalable solution for efficient streaming 3D applications.

\end{itemize}

\section{Related Work}

\subsection{Learning-Based 3D Reconstruction}
By implicitly encoding scene priors within their learned parameters, recent neural network-based approaches have substantially advanced the robustness and generalization of 3D vision models, establishing new performance standards across a variety of tasks \cite{wang2025vggt}.
A significant milestone in this progression was DUSt3R \cite{wang2024dust3r}, which demonstrated the direct regression of view-consistent 3D pointmaps from a pair of RGB images, thereby circumventing the traditional requirement for explicit camera calibration.
Building on this foundation, CUT3R \cite{wang2025continuous} introduced a stateful recurrent framework capable of incrementally updating a unified scene representation from a sequential stream of images.
Extending this paradigm, TTT3R \cite{chen2025ttt3r} incorporates a Test-Time Training (TTT) strategy that dynamically refines memory updates by assessing the alignment confidence between previously encoded states and incoming observations, thereby enhancing generalization over extended sequences. 
A pivotal advancement was realized with VGGT \cite{wang2025vggt}, which scales this philosophy within a 1.2 billion-parameter Alternative-Attention transformer architecture. 
By jointly predicting multiple 3D attributes, VGGT achieves state-of-the-art performance across a comprehensive range of 3D vision tasks. 
Most recently, StreamVGGT \cite{zhuo2025streaming} adapts this powerful model to the demands of online streaming reconstruction. 
However, its reliance on an unbounded KV cache presents a critical bottleneck, substantially impeding its practical viability and scalability in real-world streaming applications.

% \subsection{KV Cache Compression}
% KV cache enables efficient inference by avoiding redundant recomputation of past KVs. 
% As context lengths increase, the KV cache can grow into a substantial memory bottleneck, emphasizing the need for effective compression \cite{liu2024kivi}. 
% KV pruning methods reduce memory and computation by discarding past KVs of low estimated importance, typically guided by attention scores or heuristic saliency \cite{ye2025fit,wu2024accelerating}. 
% % Such approaches reveal that much of the cached KVs are redundant, allowing significant cache reduction with little effect on model performance~\cite{ye2025fit,wu2024accelerating}.
% KV quantization methods, by representing high-precision floating-point tensors in low-bit formats, enable compact KV cache storage~\cite{liu2024kivi, su2025rotatekv,su2025kvsink}. 
% Nevertheless, existing methods are designed for LLMs, and the compression of KV caches in 3D vision models remains largely unexplored.

\subsection{KV Cache Compression}
The KV cache facilitates efficient autoregressive inference by storing previously computed Key and Value representations, thereby circumventing the need for redundant recomputation. 
However, as sequence lengths grow, the KV cache rapidly becomes a critical bottleneck in terms of memory footprint, motivating a substantial body of research on cache compression techniques~\cite{li2024survey,su2025kvsink}. 
Existing approaches to KV cache compression can be broadly classified into two main categories: pruning-based and quantization-based methods. 
KV pruning seeks to reduce memory and computational overhead by selectively discarding past KV entries deemed to have low importance. 
Importance is typically estimated using criteria such as attention scores, token saliency, or other heuristics~\cite{ye2025fit,wu2024accelerating}. 
In parallel, KV quantization compresses the cache by representing high-precision Key and Value tensors in low-bit formats, thereby reducing storage requirements and memory access costs~\cite{liu2024kivi,su2025rotatekv,su2025akvq,hooper2024kvquant}. 
Despite the significant progress made in these areas, existing KV cache compression techniques have primarily been developed for LLMs that operate on textual data. 
In contrast, KV caches in 3D vision models exhibit spatial and temporal redundancies, presenting greater potential for compression.
However, effectively compressing KV caches in the context of 3D vision tasks remains an open and underexplored research challenge.

\section{Methodology}
\begin{figure*}[t]
 \vspace{-5mm}
\centering
%\vspace{-7mm}
\includegraphics[width=1\linewidth]{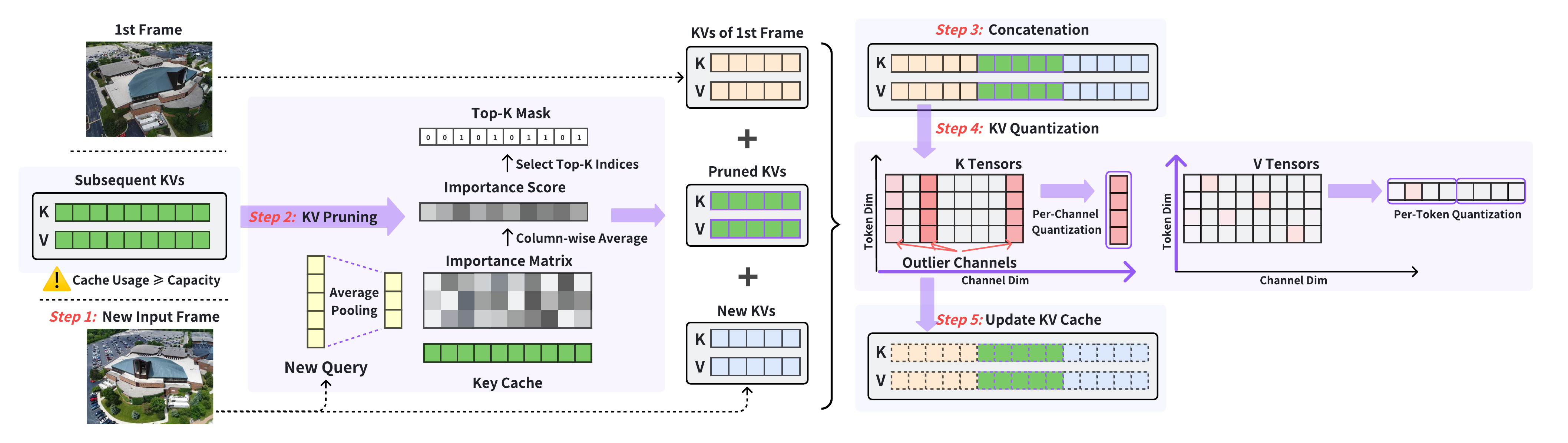}
% \vspace{-2mm}
\caption{ Overview of XStreamVGGT.
Upon receiving a new input frame (Step~1), Queries from the global attention layer are aggregated via average pooling to form a compact representation, which is then matched against the Key to estimate token importance (Step~2). 
Guided by these Key-derived importance scores, low-importance historical KV pairs are selectively pruned, while KVs from the first frame are explicitly retained to preserve geometric consistency. 
The remaining high-importance KVs are concatenated with the first-frame KVs and the newly generated KVs from the current frame (Step~3). 
Following pruning, the KV cache is further compressed using dimension-adaptive quantization, employing per-channel Key quantization and per-token Value quantization to reduce the impact of outlier channels on quantization accuracy (Step~4).
This results in a compact KV cache for efficient subsequent updates (Step~5).
}
\vspace{-5mm}
\label{fig:method}
\end{figure*}

In this section, we present XStreamVGGT, with an overview illustrated in Figure~\ref{fig:method}. 
We begin by reviewing the preliminaries of StreamVGGT in Section \ref{Preliminaries}, then present our approach to eliminating multi-frame redundancy through KV cache pruning in Section \ref{pruning}. 
In Section \ref{quantization}, we analyze the distributional properties of KV tensors in StreamVGGT, which motivate the development of an effective dimension-adaptive KV quantization scheme.
\subsection{Preliminaries}
\label{Preliminaries}
StreamVGGT is a streaming 4D visual geometry transformer that processes video frames in an online manner, producing per-frame geometry outputs without the need to reprocess the entire history. 
Given an incoming RGB frame \( I_t \in \mathbb{R}^{3 \times H \times W} \) at time step \( t \), the model first converts it into a sequence of visual tokens \( F_t \in \mathbb{R}^{N \times C} \) using a patch embedding network, where \( N \) is the number of image patches and \( C \) is the embedding dimension.
In addition to the patch tokens, StreamVGGT prepends a camera token \( g_t \in \mathbb{R}^{1 \times C} \) to encode global camera-related information, which is later decoded by task-specific heads to predict camera parameters. 
The model also includes \( R \) register tokens \( r_t \in \mathbb{R}^{R \times C} \), which act as auxiliary latent slots to absorb redundant attention responses during transformer processing.
Thus, the input token sequence for frame \( t \) consists of the camera token, register tokens, and patch tokens, with a total length of \( 1 + R + N \). 
This token sequence is fed into a spatio-temporal transformer encoder with \( L \) layers, adopting an Alternating-Attention design \cite{wang2025vggt}. 
Each layer first applies frame-wise spatial self-attention to model intra-frame structure, followed by temporal causal attention to aggregate information from past frames under a strict causal constraint.

In each transformer layer $\ell$, the temporal attention module maintains a KV cache that stores token-level representations from all previous frames:
\begin{equation}
\mathcal{C}^{(\ell)}_{t-1} = \{ K^{(\ell)}_{1:t-1}, V^{(\ell)}_{1:t-1} \},
\end{equation}
where $K^{(\ell)}_{\tau}, V^{(\ell)}_{\tau} \in \mathbb{R}^{(1+R+N)\times C}$ denote the Key and Value tensors corresponding to all tokens of frame $\tau$. 
For the current frame, only the Query, Key, and Value tensors $Q^{(\ell)}_t$, $K^{(\ell)}_t$, and $V^{(\ell)}_t$ are newly computed. 
Temporal attention at layer $\ell$ is computed as:
\begin{equation}
\mathrm{Attn}\bigl(
Q^{(\ell)}_t, \;
[ K^{(\ell)}_{1:t-1}, K^{(\ell)}_t ], \;
[ V^{(\ell)}_{1:t-1}, V^{(\ell)}_t ]
\bigr),
\end{equation}
with a causal mask to prevent access to future frames. 
After the attention computation, $K^{(\ell)}_t$ and $V^{(\ell)}_t$ are appended to the cache for use in subsequent time steps.
The outputs of the final transformer layer are then passed to lightweight, task-specific heads to predict per-frame geometry signals, such as camera parameters, dense point maps, and others.

During inference, the size of the Query tensor $Q^{(\ell)}_t$ remains constant over time, as it depends solely on the tokens of the current frame. 
In contrast, the Key and Value tensors grow linearly with the number of processed frames due to the accumulation of cached representations. Consequently, the memory cost of temporal attention at each layer increases linearly with the sequence length, making long-duration streaming inference progressively more expensive.

\subsection{Eliminating Multi-Frame Redundancy through KV Cache Pruning}
\label{pruning}
\begin{figure*}[t]
 \vspace{-5mm}
\centering
    \begin{subfigure}{0.33\linewidth}
        \centering
    \includegraphics[width=\linewidth]{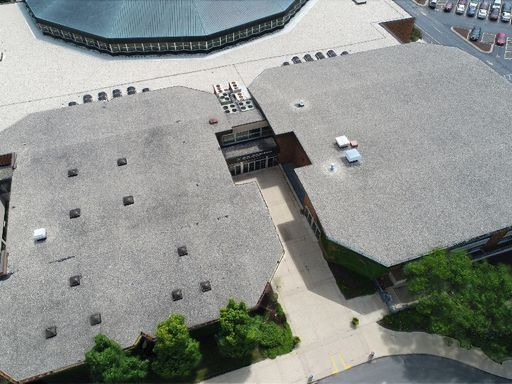}
    \caption{The input query frame.}
    \label{fig:query-frame}
    \end{subfigure}
    \begin{subfigure}{0.664\linewidth}
    \centering
    \includegraphics[width=\linewidth]{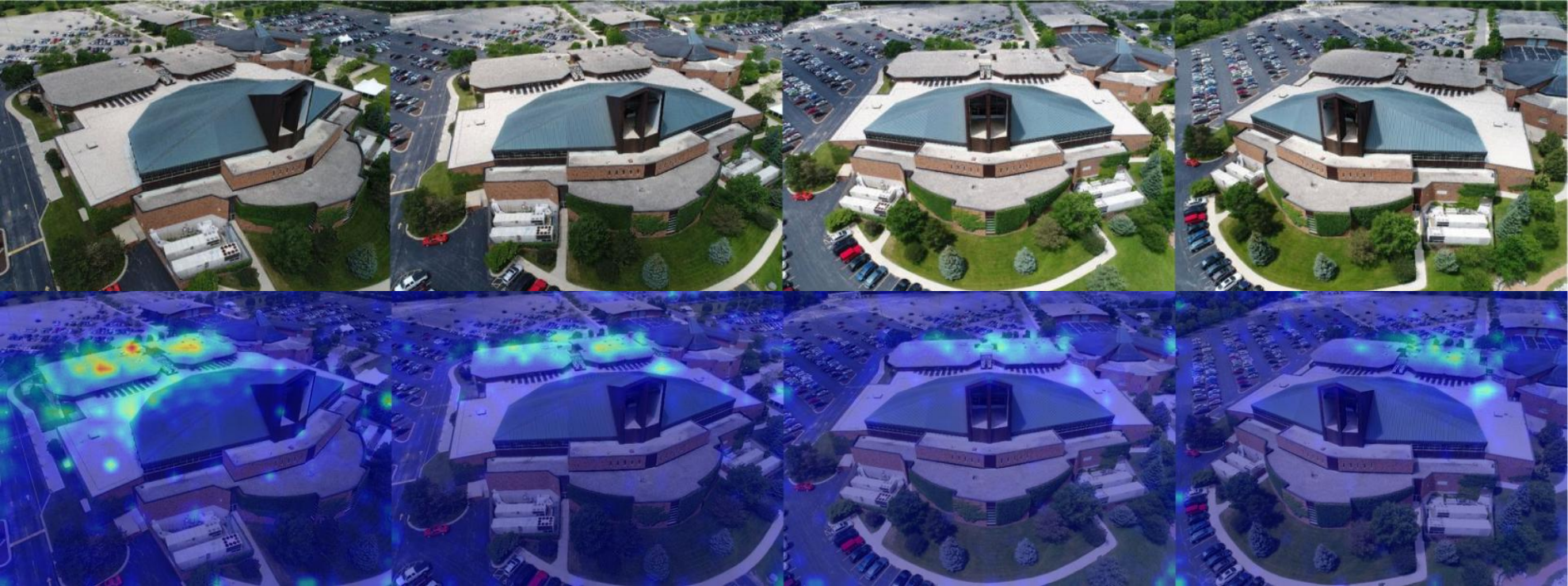}
    \caption{Attention heatmaps of previous input frames.}
    \label{fig:frame1234}
    \end{subfigure}
\caption{Attention sparsity analysis. The visualization shows attention heatmaps from Layer 14 of StreamVGGT. The visualization of attention heatmaps reveals that attention weights are predominantly concentrated on Query-relevant regions. 
In contrast, other areas exhibit significantly lower attention, indicating substantial redundancy in the feature representation.}

\label{fig:redundency}
\end{figure*}

Despite the shared use of KV caching, StreamVGGT differs fundamentally from autoregressive LLMs. 
In LLMs, KV caches are formed from text tokens carrying rich semantic context, whereas in StreamVGGT they originate from visual tokens extracted from video frames. 
Unlike textual tokens, vision tokens exhibit substantial redundancy due to both intra-frame spatial correlations and inter-frame temporal consistency. 
As illustrated in Figure \ref{fig:redundency}, this redundancy manifests as sparse attention patterns across multi-frame inputs, suggesting significant opportunities for cache compression. 

We propose a query-guided KV cache pruning mechanism to eliminate multi-frame redundancy while retaining the most informative historical tokens within a fixed cache length \( \mathcal{L}_{\text{max}} \). 
The pruning operation is applied independently at each transformer layer. Notably, since the spatio-temporal encoder employs an Alternating-Attention design, only the temporal global attention module maintains a KV cache and is thus subject to pruning.
We begin by computing the similarity between the pooled Query tokens of the current frame and each Key token to assess token importance. 
This approach strikes a balance between using attention scores as a metric for token importance and efficiently computing the attention kernel. 
Attention scores have been widely used in previous studies to identify token importance \cite{ye2025fit,wu2024accelerating}. 
However, the attention calculation in StreamVGGT is designed to be compatible with highly optimized, efficient attention kernels (such as FlashAttention \cite{dao2023flashattention,zhang2026snapmla}), which do not provide intermediate computation results, including attention scores.
While repeatedly computing the \( QK \) attention scores yields accurate attention scores, it incurs excessive computational cost. By leveraging pooled Query groups for importance identification, we efficiently compute token importance while ensuring compatibility with optimized attention kernels. 

% TBD
Specifically, given the Query \( Q^{(\ell)}_t \in \mathbb{R}^{(1+R+N) \times C} \) of the current frame at layer \( \ell \), we first separate the special tokens (camera token and register tokens), denoted as \( Q^{(\ell)}_{t,\text{special}} \), from the ordinary patch tokens \( Q^{(\ell)}_{t,\text{normal}} \). 
The patch tokens are grouped into fixed-size groups of size \( g \).
The pooled Query is then formed as:
\begin{equation}
Q^{(\ell)}_{t,\text{pooled}} =
\text{concat}\Big(
Q^{(\ell)}_{t,\text{special}},
\; \text{GroupAvg}(Q^{(\ell)}_{t,\text{normal}}, g)
\Big).
\end{equation}
We further average the pooled Query across all attention heads to obtain:
\begin{equation}
\bar{Q}^{(\ell)}_t =
\frac{1}{H} \sum_{h=1}^{H} Q^{(\ell)}_{t,\text{pooled}}[h]
\in \mathbb{R}^{N_{\text{pooled}} \times C},
\end{equation}
where \( N_{\text{pooled}} \) denotes the number of tokens after grouping. 
The grouping size \( g \) is a fixed hyperparameter.

Pruning is triggered immediately after the temporal global attention at layer \( \ell \) completes, once the total cache length \( T \) exceeds the budget \( \mathcal{L}_{\text{max}} \). 
Tokens from the first and current frames are always preserved, serving as stable geometric references and up-to-date visual evidence, respectively. 
Let \( T_{\text{first}} \) and \( T_{\text{current}} \) denote the number of tokens corresponding to the first and current frames. 
The remaining middle segment is subject to pruning: \( T_{\text{prunable}} = T - T_{\text{first}} - T_{\text{current}} \).
For the prunable middle tokens, we compute a head-averaged key summary:
\begin{equation}
\bar{K}^{(\ell)}_{\text{prunable}} = 
\frac{1}{H}
\sum_{h=1}^{H}
K^{(\ell)}_{1:t-1}
\bigl[h,\; T_{\text{first}} : T - T_{\text{current}} \bigr]
\in \mathbb{R}^{T_{\text{prunable}} \times C}.
\end{equation}
All token types in the middle frames, including camera, register, and patch tokens, are treated uniformly during pruning.
Token importance scores are computed via the inner product between the pooled queries and the prunable keys:
\begin{equation}
S^{(\ell)}_{\text{matrix}} = 
\bar{Q}^{(\ell)}_t
\left(\bar{K}^{(\ell)}_{\text{prunable}}\right)^{\!\top}
\in
\mathbb{R}^{N_{\text{pooled}} \times T_{\text{prunable}}},
\end{equation}
followed by averaging along the Query dimension:
\begin{equation}
S^{(\ell)} = 
\frac{1}{N_{\text{pooled}}}
\sum_{i=1}^{N_{\text{pooled}}}
S^{(\ell)}_{\text{matrix}}[i, :]
\in
\mathbb{R}^{T_{\text{prunable}}}.
\end{equation}

Based on the resulting importance scores, the top-\(k\) tokens are selected from the prunable middle region.
% , where
% \begin{equation}
% k = \max\bigl(0,\; \mathcal{L}_{\text{max}} - T_{\text{first}} - T_{\text{current}} \bigr).
% \end{equation}
Let \( \mathcal{I}_{\text{middle}} \) denote the indices of the selected tokens. The final set of retained tokens is then given by
\begin{equation}
\mathcal{I}_{\text{keep}} = 
\{1, \dots, T_{\text{first}}\}
\;\cup\;
\mathcal{I}_{\text{middle}}
\;\cup\;
\{T - T_{\text{current}} + 1, \dots, T\}.
\end{equation}
The same indices are synchronously applied to both the Key and Value tensors to maintain consistency.
With the proposed pruning strategy, the cache size increases linearly only until reaching the budget \( \mathcal{L}_{\text{max}} \), after which it remains fixed. 
Consequently, the per-frame temporal attention cost transitions from linear growth to a constant upper bound.

% TBD
\begin{figure}[t]
 \vspace{-5mm}
    \centering  

    \begin{subfigure}{0.245\linewidth}
        \centering
    \includegraphics[width=\linewidth]{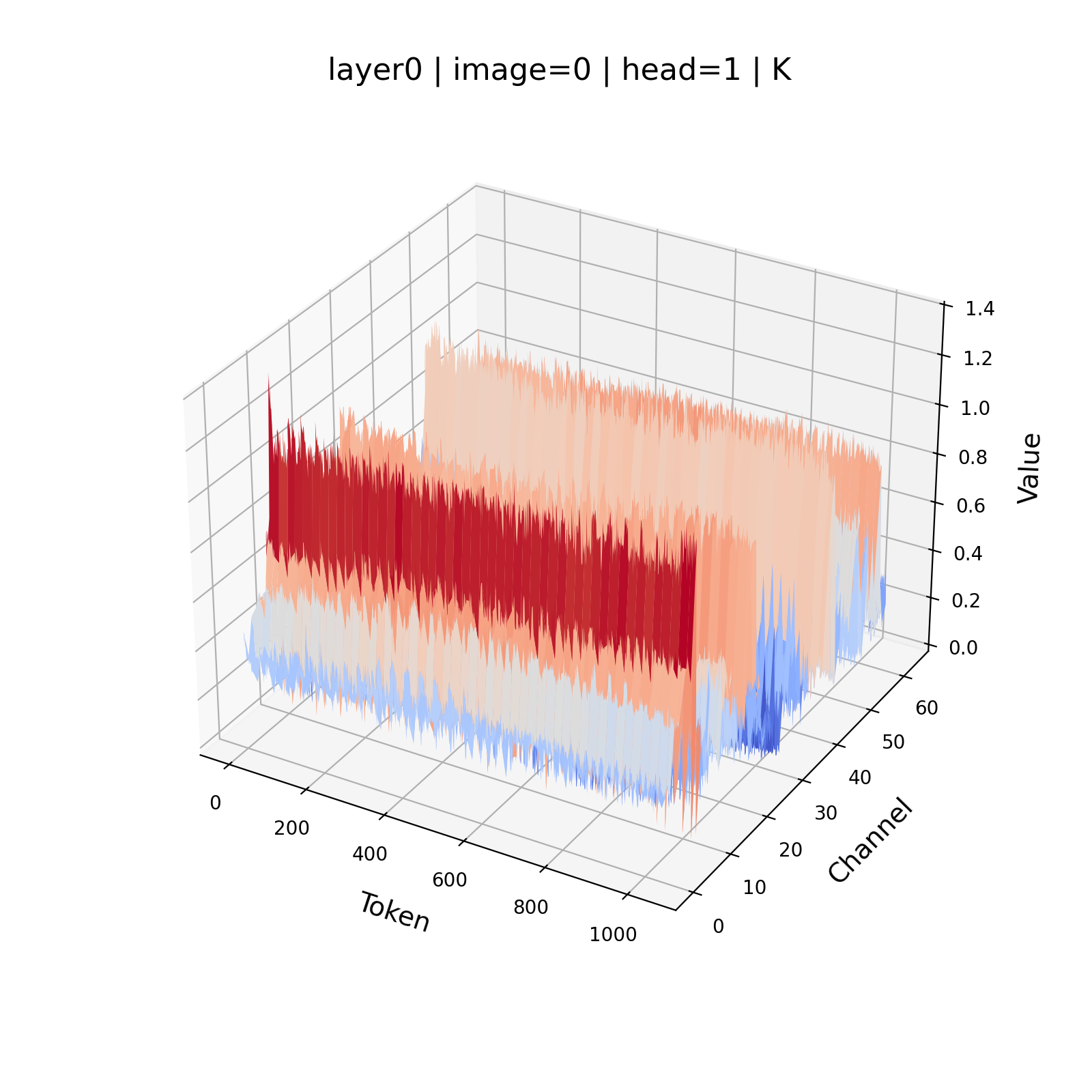}
    \caption{Key tensor.}
    \end{subfigure}
    \begin{subfigure}{0.245\linewidth}
        \centering
    \includegraphics[width=\linewidth]{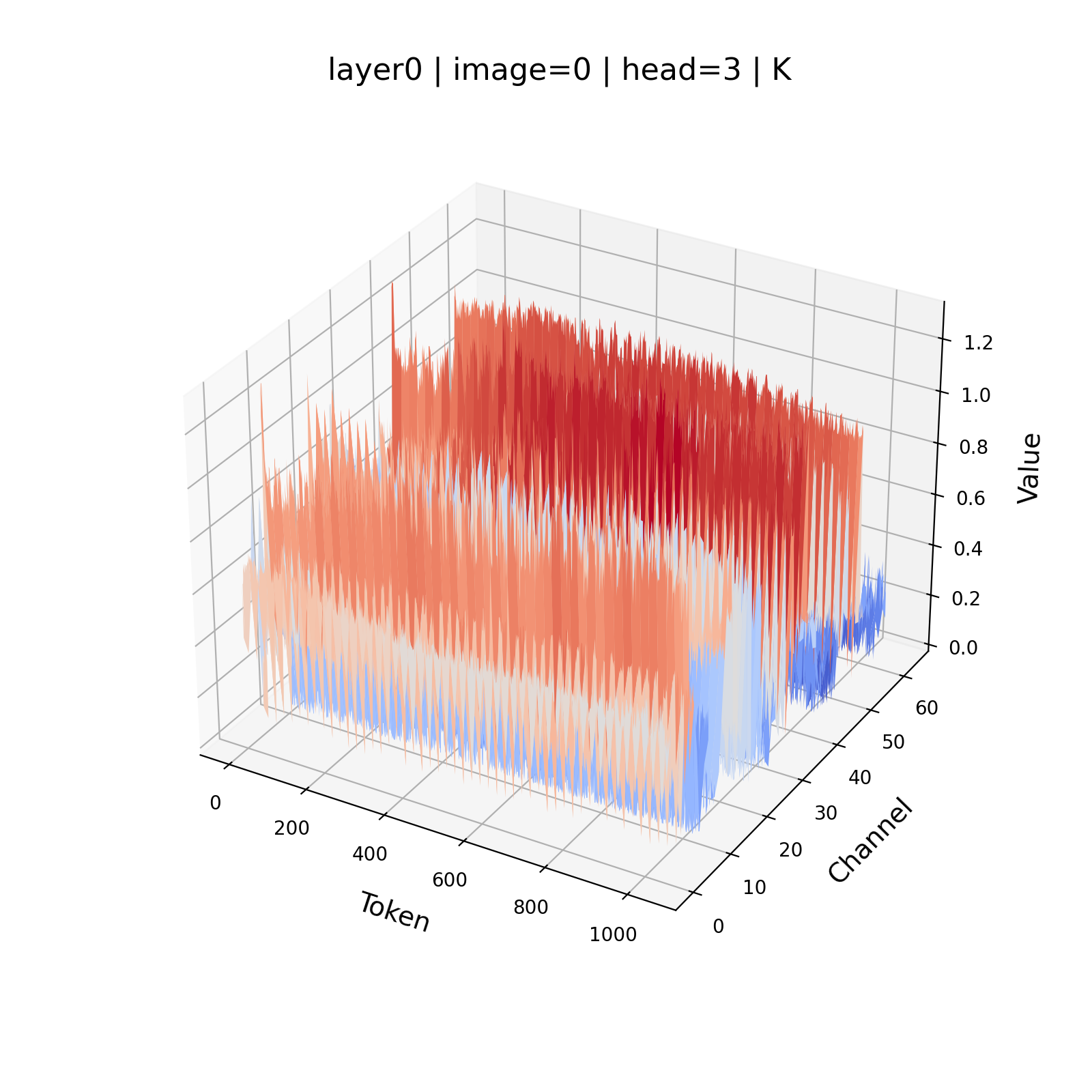}
    \caption{Key tensor.}
    \end{subfigure}
    \begin{subfigure}{0.245\linewidth}
        \centering
    \includegraphics[width=\linewidth]{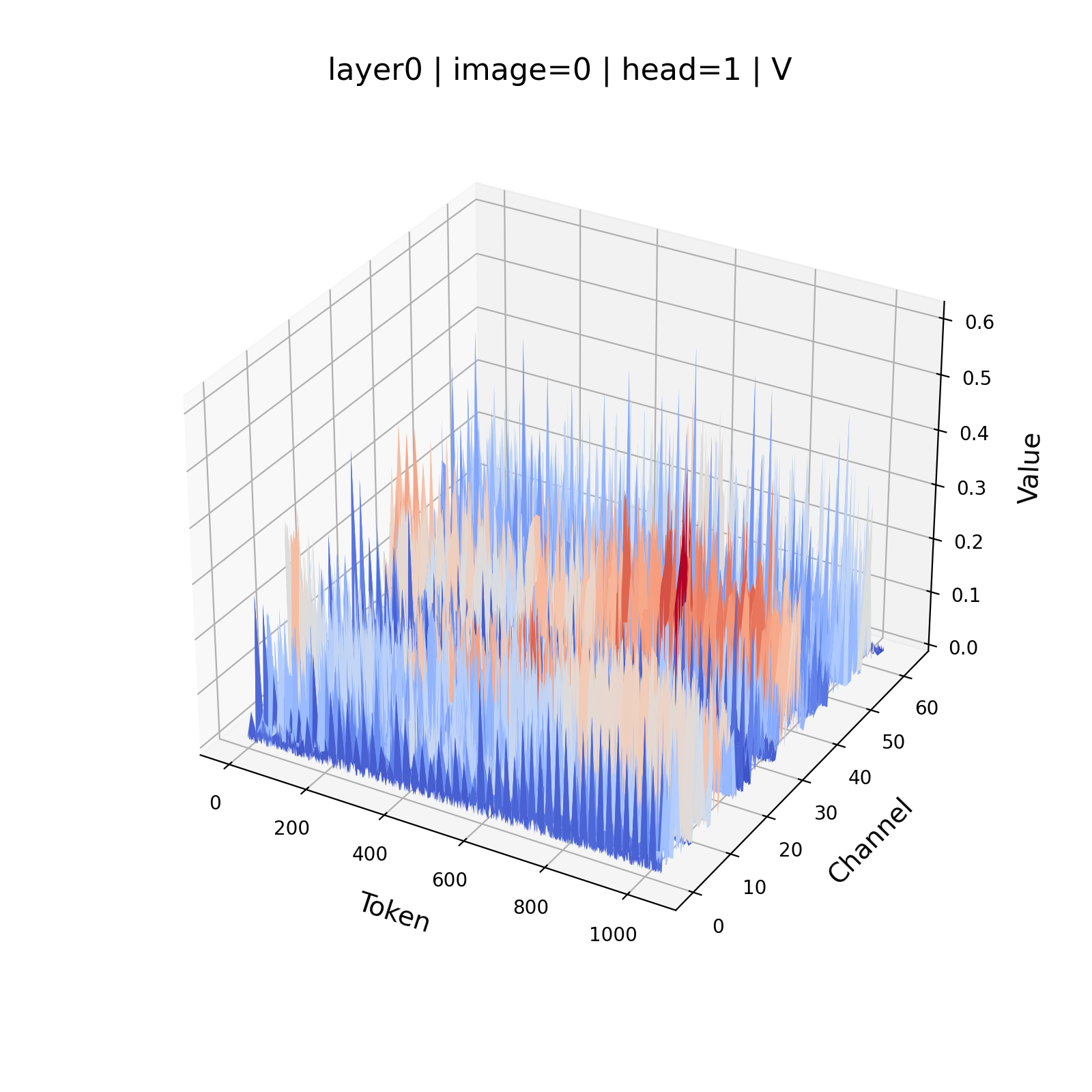}
    \caption{Value tensor.}
    \end{subfigure}
    \begin{subfigure}{0.245\linewidth}
        \centering
    \includegraphics[width=\linewidth]{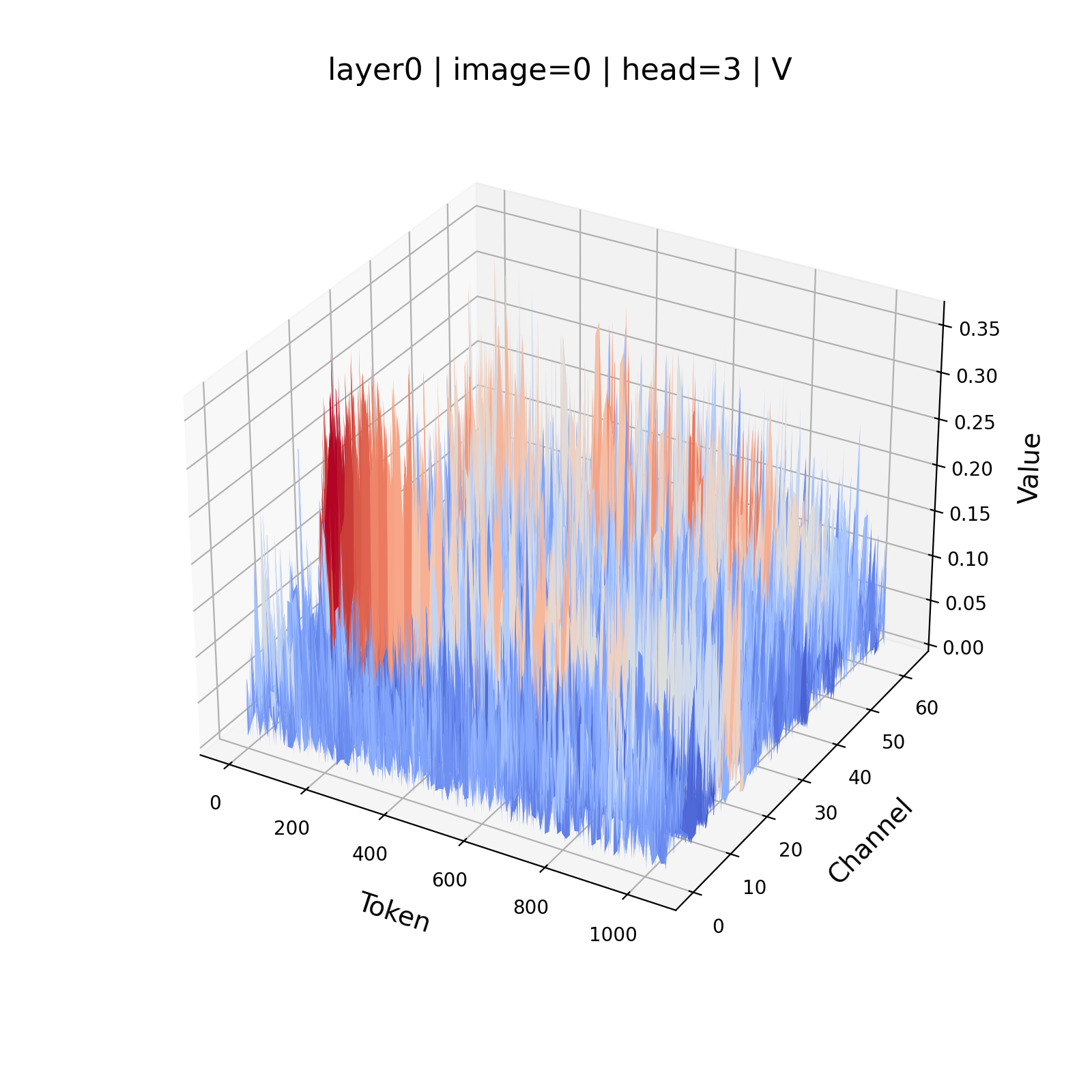}
    \caption{Value tensor.}
    \end{subfigure}
\caption{Magnitude distributions of the Key and Value. 
The Key demonstrates significant channel-wise outliers, with a small subset of channels exhibiting magnitudes substantially larger than the others. In contrast, the distribution of the Value is more uniform, with no prominent outlier behavior.}
    % %\vspace{-3mm}
\label{fig:kv tensor}
\end{figure}
\subsection{Dimension-Adaptive KV Quantization Based on KV Distribution Characteristics}
\label{quantization}
To achieve extreme compression of the KV cache, we leverage the inherent distribution patterns of KV tensors and propose a dimension-adaptive KV quantization scheme. 
In this work, we adopt the widely used asymmetric uniform quantization scheme \cite{liu2024kivi}, which is parameterized by a scale factor $s$, a zero-point $z$, and a bit-width $b$.
Given a floating-point tensor $x \in \mathbb{R}^{d}$, its quantized representation $\hat{x}$ is computed as
\begin{equation}
%\vspace{-2mm}
\hat{x}
=
\operatorname{clamp}
\left(
\left\lfloor \frac{x}{s} \right\rceil + z,\;
0,\;
2^{b}-1
\right),
\end{equation}
with the scale factor and zero-point defined as
\begin{equation}
%\vspace{-2mm}
s
=
\frac{x_{\max} - x_{\min}}{2^{b} - 1},
\qquad
z
=
\left\lfloor -\frac{x_{\min}}{s} \right\rceil.
\end{equation}
Here, $\lfloor \cdot \rceil$ denotes rounding to the nearest integer, and $\operatorname{clamp}(\cdot)$ restricts values to the valid range.
The scale factor $s$ determines the quantization step size, while the zero-point $z$ ensures that the zero is mapped to the quantized range.

% TBD
\begin{table}[t]
\vspace{-3mm}
\centering
\caption{Analysis of quantization errors for per-token and per-channel schemes. The errors are computed based on the Mean Squared Error (MSE) metric with a group size of 64.}
\resizebox{0.8\textwidth}{!}{%
\label{tab:quantization-errors}
\centering
\begin{tabular}{ccc|c}
\hline
 & Bits & Per-Token Quantization & Per-Channel Quantization \\ \hline
\multirow{2}{*}{Keys} & INT4 & $2.007 \times 10^{-3}$ & $3.635 \times 10^{-4}$ \\
 & INT2 & $5.183 \times 10^{-2}$ & $9.181 \times 10^{-3}$ \\ \hline
\multirow{2}{*}{Values} & INT4 & $5.035 \times 10^{-4}$ & $4.704 \times 10^{-4}$ \\
 & INT2 & $2.641 \times 10^{-2}$ & $2.190 \times 10^{-2}$ \\ \hline
\end{tabular}%
% \vspace{-3mm}
}
\end{table}%

After establishing the basic quantization strategy, we next determine the appropriate quantization granularity (i.e., per-tensor, per-token, or per-channel).
A principled understanding of the distributional properties of KV tensors in StreamVGGT is crucial for achieving robust and accurate quantization.
We conducted an extensive analysis of the KV distribution in StreamVGGT, revealing the distinct distribution patterns of Keys and Values within 3D reconstruction transformer models.
As shown in Figure \ref{fig:kv tensor}, the Key tensors exhibit significant channel-wise outliers, a behavior that is much less pronounced in the Value tensors. 
When applying standard per-tensor or per-token quantization, these outliers dominate the dynamic range, leading to inflated quantization scales and a marked reduction in effective precision, which, in turn, causes substantial performance degradation \cite{nagel2021white}.

Guided by these observations, we propose a per-channel Key and per-token Value quantization scheme that explicitly addresses channel-wise outliers. 
As presented in Table \ref{tab:quantization-errors}, our analysis confirms that this scheme significantly mitigates quantization error.
Unlike LLMs, where the KV cache typically expands by a single token at each decoding step, StreamVGGT processes inputs in a frame-wise manner. 
This frame-wise processing results in a rapid expansion of the KV cache, adding a large number of patch tokens per time step. 
As a result, StreamVGGT is inherently well-suited for per-channel Key quantization schemes that span multiple tokens.
As illustrated in Figure~\ref{fig:method}, we tightly couple quantization with pruning.
Specifically, quantization is applied to the final KV cache, $\tilde{K}^{(\ell)}_{1:t-1}$ and $\tilde{V}^{(\ell)}_{1:t-1}$, which comprises both the pruned historical KVs and the preserved KVs from the first and current frames.
The quantization is defined as
\begin{equation}
%\vspace{-2mm}
\hat{K}_{c} = \mathcal{Q}_{c}\!\left(\tilde{K}_{c}; \, s^{K}_{c}, \, z^{K}_{c}\right),
\qquad
\hat{V}_{t} = \mathcal{Q}_{t}\!\left(\tilde{V}_{t}; \, s^{V}_{t}, \, z^{V}_{t}\right),
\end{equation}
where $c$ and $t$ denote channel and token indices, respectively, 
and $s^{K}_{c}, z^{K}_{c}$ (for Keys) and $s^{V}_{t}, z^{V}_{t}$ (for Values) are the corresponding scale and zero-point parameters.
Here, $\mathcal{Q}_{c}(\cdot)$ and $\mathcal{Q}_{t}(\cdot)$ denote asymmetric uniform quantization along the channel and token dimensions, respectively.
During attention computation, the quantized tensors are dequantized as
\begin{equation}
%\vspace{-2mm}
\tilde{K}_{c} = \mathcal{DQ}_{c}\!\left(\hat{K}_{c}; \, s^{K}_{c}, \, z^{K}_{c}\right),
\qquad
\tilde{V}_{t} = \mathcal{DQ}_{t}\!\left(\hat{V}_{t}; \, s^{V}_{t}, \, z^{V}_{t}\right),
\end{equation}
where $\mathcal{DQ}_{c}(\cdot)$ and $\mathcal{DQ}_{t}(\cdot)$ denote the dequantization operators.

\begin{table}[t]
 \vspace{-5mm}
\centering
\caption{3D reconstruction comparison on NRGBD dataset.}
\resizebox{1\textwidth}{!}{%
\begin{tabular}{@{}ccccccccc@{}}
\toprule
 &  &  & \multicolumn{6}{c}{NRGBD} \\ \cmidrule(l){4-9} 
 &  &  & \multicolumn{2}{c}{Acc $\downarrow$} & \multicolumn{2}{c}{Comp $\downarrow$} & \multicolumn{2}{c}{NC $\uparrow$} \\
Method & \multicolumn{2}{c}{\textit{Inference Type}} & Mean & Med. & Mean & Med. & Mean & Med. \\ \midrule
VGGT & \textit{Offline} & \textit{Unbounded Memory} & 0.079 & 0.020 & 0.083 & 0.023 & 0.909 & 0.989 \\
StreamVGGT & \textit{Online} & \textit{Unbounded Memory} & 0.085 & 0.044 & 0.079 & 0.038 & 0.862 & 0.986 \\
\textbf{XStreamVGGT (Ours)} & \textit{\textbf{Online}} & \textit{\textbf{Bounded Memory}} & \textbf{0.085} & \textbf{0.049} & \textbf{0.075} & \textbf{0.038} & \textbf{0.850} & \textbf{0.986} \\ \bottomrule
\end{tabular}%
}
  \label{tab:mv_recon_1}%
\end{table}%

\begin{table}[t]
\centering
\caption{3D reconstruction comparison on 7-Scenes dataset.}
\resizebox{1\textwidth}{!}{%
\begin{tabular}{@{}ccccccccc@{}}
\toprule
 &  &  & \multicolumn{6}{c}{7 Scenes} \\ \cmidrule(l){4-9} 
 &  &  & \multicolumn{2}{c}{Acc $\downarrow$} & \multicolumn{2}{c}{Comp $\downarrow$} & \multicolumn{2}{c}{NC $\uparrow$} \\
Method & \multicolumn{2}{c}{\textit{Inference Type}} & Mean & Med. & Mean & Med. & Mean & Med. \\ \midrule
VGGT & \textit{Offline} & \textit{Unbounded Memory} & 0.085 & 0.038 & 0.089 & 0.039 & 0.786 & 0.889 \\
StreamVGGT & \textit{Online} & \textit{Unbounded Memory} & 0.132 & 0.058 & 0.116 & 0.042 & 0.749 & 0.863 \\
\textbf{XStreamVGGT (Ours)} & \textit{\textbf{Online}} & \textit{\textbf{Bounded Memory}} & \textbf{0.142} & \textbf{0.068} & \textbf{0.125} & \textbf{0.048} & \textbf{0.734} & \textbf{0.848} \\ \bottomrule
\end{tabular}%
}
  \label{tab:mv_recon_2}%
\end{table}%

\begin{table}[t]
\centering
 \vspace{-5mm}
\caption{Camera pose estimation.}
\resizebox{\textwidth}{!}{%
\begin{tabular}{@{}ccccccccc@{}}
\toprule
 &  &  & \multicolumn{3}{c}{TUM} & \multicolumn{3}{c}{ScanNet} \\ \cmidrule(l){4-9} 
Method & \multicolumn{2}{c}{\textit{Inference Type}} & ATE $\downarrow$ & RPE trans $\downarrow$ & \multicolumn{1}{c|}{RPE rot $\downarrow$} & ATE $\downarrow$ & RPE trans $\downarrow$ & RPE rot $\downarrow$ \\ \midrule
VGGT & \textit{Offline} & \textit{Unbounded Memory} & 0.053 & 0.032 & \multicolumn{1}{c|}{3.209} & 0.157 & 0.056 & 3.635 \\
StreamVGGT & \textit{Online} & \textit{Unbounded Memory} & 0.062 & 0.033 & \multicolumn{1}{c|}{3.208} & 0.160 & 0.057 & 3.688 \\
\textbf{XStreamVGGT (Ours)} & \textit{\textbf{Online}} & \textit{\textbf{Bounded Memory}} & \textbf{0.068} & \textbf{0.035} & \multicolumn{1}{c|}{\textbf{3.184}} & \textbf{0.171} & \textbf{0.061} & \textbf{3.837} \\ \bottomrule
\end{tabular}%
}
  \label{tab:camera_pose}%
\end{table}%
\section{Experiments}

\subsection{Experimental Details}
We perform a comprehensive evaluation of XStreamVGGT, encompassing benchmark performance, efficiency analysis, ablation studies, and qualitative results. 
XStreamVGGT is first assessed on various 3D tasks, including 3D reconstruction, camera pose estimation and depth estimation. 
Following the Point3R protocol \cite{wu2025point3r}, input images are processed with variable aspect ratios and resized to ensure the maximum edge length does not exceed 518 pixels.
For pruning, the pooling size is set to 16, and the cache length is set to 2K, ensuring that at least the first and current frames are preserved. 
KV quantization is performed using KIVI with INT4 and a group size of 64 \cite{liu2024kivi}.
Remarkably, even with a cache length of only 2K and additional quantization, XStreamVGGT achieves exceptional efficiency while maintaining robust performance, with negligible loss across most tasks compared to the more expensive full-KV-length StreamVGGT \cite{zhuo2025streaming}.

\begin{table}[t]
\centering
\caption{Monocular depth estimation.}
\resizebox{1\textwidth}{!}{%
\begin{tabular}{@{}ccccccccc@{}}
\toprule
 &  &  & \multicolumn{2}{c}{Sintel} & \multicolumn{2}{c}{Bonn} & \multicolumn{2}{c}{KITTI} \\ \cmidrule(l){4-9} 
Method & \multicolumn{2}{c}{Inference Type} & Abs Rel $\downarrow$ & \multicolumn{1}{c|}{${\delta}$\textless{}1.25 $\uparrow$} & Abs Rel $\downarrow$ & \multicolumn{1}{c|}{${\delta}$\textless{}1.25 $\uparrow$} & Abs Rel $\downarrow$ & ${\delta}$\textless{}1.25 $\uparrow$ \\ \midrule
VGGT & \textit{Offline} & \textit{Unbounded Memory} & 0.274 & \multicolumn{1}{c|}{67.4} & 0.054 & \multicolumn{1}{c|}{97.2} & 0.071 & 94.1 \\
StreamVGGT & \textit{Online} & \textit{Unbounded Memory} & 0.254 & \multicolumn{1}{c|}{68.5} & 0.052 & \multicolumn{1}{c|}{97.1} & 0.072 & 94.7 \\
\textbf{XStreamVGGT} & \textit{\textbf{Online}} & \textit{\textbf{Bounded Memory}} & \textbf{0.254} & \multicolumn{1}{c|}{\textbf{68.5}} & \textbf{0.052} & \multicolumn{1}{c|}{\textbf{97.1}} & \textbf{0.072} & \textbf{94.7} \\ \bottomrule
\end{tabular}%
}
  \label{tab:monodepth}%
\end{table}%

\begin{table}[t]
 \vspace{-5mm}
\centering
\caption{Video depth evaluation.}
\resizebox{1\textwidth}{!}{%
\begin{tabular}{@{}ccccccccc@{}}
\toprule
 &  & \multicolumn{1}{l}{} & \multicolumn{2}{c}{Sintel} & \multicolumn{2}{c}{Bonn} & \multicolumn{2}{c}{KITTI} \\ \cmidrule(l){4-9} 
Method & \multicolumn{2}{c}{Inference Type} & Abs Rel $\downarrow$ & \multicolumn{1}{c|}{${\delta}$\textless{}1.25 $\uparrow$} & Abs Rel $\downarrow$ & \multicolumn{1}{c|}{${\delta}$\textless{}1.25 $\uparrow$} & Abs Rel $\downarrow$ & ${\delta}$\textless{}1.25 $\uparrow$ \\ \midrule
VGGT & \textit{Offline} & \textit{Unbounded Memory} & 0.301 & \multicolumn{1}{c|}{68.3} & 0.057 & \multicolumn{1}{c|}{96.8} & 0.061 & 97.0 \\
StreamVGGT & \textit{Online} & \textit{Unbounded Memory} & 0.328 & \multicolumn{1}{c|}{65.8} & 0.058 & \multicolumn{1}{c|}{97.2} & 0.094 & 94.4 \\
\textbf{XStreamVGGT} & \textit{\textbf{Online}} & \textit{\textbf{Bounded Memory}} & \textbf{0.341} & \multicolumn{1}{c|}{\textbf{61.9}} & \textbf{0.073} & \multicolumn{1}{c|}{\textbf{94.3}} & \textbf{0.098} & \textbf{94.3} \\ \bottomrule
\end{tabular}%
}
  \label{tab:video_depth}%
\end{table}%

\subsection{3D Reconstruction}
The 3D reconstruction experiments were conducted on the 7-Scenes \cite{shotton2013scene} and NRGBD \cite{azinovic2022neural} datasets by measuring the discrepancies between predictions and the corresponding ground-truth point clouds.  
We adopt Accuracy (Acc), Completion (Comp), and Normal Consistency (NC) as evaluation metrics.
Although XStreamVGGT uses bounded memory, it effectively maintains performance across various metrics.
As shown in Table \ref{tab:mv_recon_1}, on the NRGBD dataset, XStreamVGGT maintains stable performance across several metrics, with only a slight reduction in the others.
NC sees only a slight drop of 1.4\% (Mean) and 0.3\% (Med) compared to StreamVGGT, demonstrating efficient memory management with minimal trade-offs.
As shown in Table \ref{tab:mv_recon_2}, XStreamVGGT demonstrates strong geometric fidelity on the 7-Scenes dataset, achieving a mean NC score of 0.734, which represents only a 2\% drop compared to StreamVGGT (0.749).

\subsection{Camera Pose Estimation}
We evaluate camera pose estimation on the TUM Dynamics \cite{sturm2012benchmark} and ScanNet \cite{dai2017scannet} datasets, truncating all sequences to 90 frames.
% Note that both the Sintel and TUM Dynamics datasets contain substantial dynamic objects, which makes them particularly challenging for traditional SfM systems.
We report Absolute Translation Error (ATE), Relative Translation Error (RPE\textsubscript{trans}), and Relative Rotation Error (RPE\textsubscript{rot}) as evaluation metrics.
As shown in Table \ref{tab:camera_pose}, XStreamVGGT achieves nearly lossless performance on the Sintel dataset, exhibiting only a minimal increase in ATE (0.006) and translational RPE (0.002) compared with StreamVGGT. The change in rotational RPE is even smaller, with an increase of only 0.025 (approximately 0.8\%).
These minor increases highlight XStreamVGGT's ability to efficiently manage memory while maintaining nearly identical performance across key metrics.

\begin{table}[t]
  \centering
\caption{Ablation study of the pruning and quantization processes in video depth estimation.}
\resizebox{\textwidth}{!}{%
\begin{tabular}{@{}cccccccccc@{}}
\toprule
 &  &  &  & \multicolumn{2}{c}{Sintel} & \multicolumn{2}{c}{Bonn} & \multicolumn{2}{c}{KITTI} \\ \cmidrule(l){5-10} 
Method & Cache Length & Pruning & Quantization & Abs Rel $\downarrow$ & \multicolumn{1}{c|}{${\delta}$\textless{}1.25 $\uparrow$} & Abs Rel $\downarrow$ & \multicolumn{1}{c|}{${\delta}$\textless{}1.25 $\uparrow$} & Abs Rel $\downarrow$ & ${\delta}$\textless{}1.25 $\uparrow$ \\ \midrule
StreamVGGT & Unbounded & w/o & w/o & 0.358 & \multicolumn{1}{c|}{61.1} & 0.080 & \multicolumn{1}{c|}{96.5} & 0.198 & 69.0 \\
XStreamVGGT & 2K & w/ & w/o & 0.343 & \multicolumn{1}{c|}{61.4} & 0.067 & \multicolumn{1}{c|}{94.9} & 0.224 & 58.0 \\
\textbf{XStreamVGGT} & \textbf{2K} & \textbf{w/} & \textbf{w/} & \textbf{0.341} & \multicolumn{1}{c|}{\textbf{61.9}} & \textbf{0.073} & \multicolumn{1}{c|}{\textbf{94.3}} & \textbf{0.206} & \textbf{63.9} \\ \bottomrule
\end{tabular}%
}
  \label{tab:ablation}%
\end{table}%

\subsection{Depth Estimation}
Following Monst3R \cite{zhang2024monst3r}, we evaluate monocular and video depth estimation performance on the Sintel \cite{butler2012naturalistic}, Bonn \cite{palazzolo2019refusion}, and KITTI \cite{geiger2013vision} datasets, which encompass a diverse range of dynamic and static, indoor and outdoor scenes. 
The evaluation metrics include Absolute Relative Error (Abs Rel) and $\delta < 1.25$ (the percentage of predicted depths within a 1.25 factor of the ground-truth depth).
For monocular depth estimation, the results in Table \ref{tab:monodepth} demonstrate that XStreamVGGT fully preserves the performance of StreamVGGT, yielding no observable degradation in any evaluation metric.
For video depth estimation, as shown in Table \ref{tab:monodepth} and~\ref{tab:video_depth}, XStreamVGGT experiences mostly negligible performance degradation with a 2K cache length. 
Although the relative drop on Bonn is more noticeable due to its exceptionally strong baseline, XStreamVGGT still achieves very high absolute performance (0.073 Abs Rel).
In contrast to StreamVGGT, which faces significant GPU memory surges and ultimately runs into Out of Memory (OOM) issues as the number of input frames increases, XStreamVGGT maintains a consistent GPU memory usage and achieves nearly 6x speedup with lossless results.

\subsection{Ablation Study}
\begin{figure}[h!]
    \centering  
    \begin{subfigure}{0.5\linewidth}
        \centering
    \includegraphics[width=\linewidth]{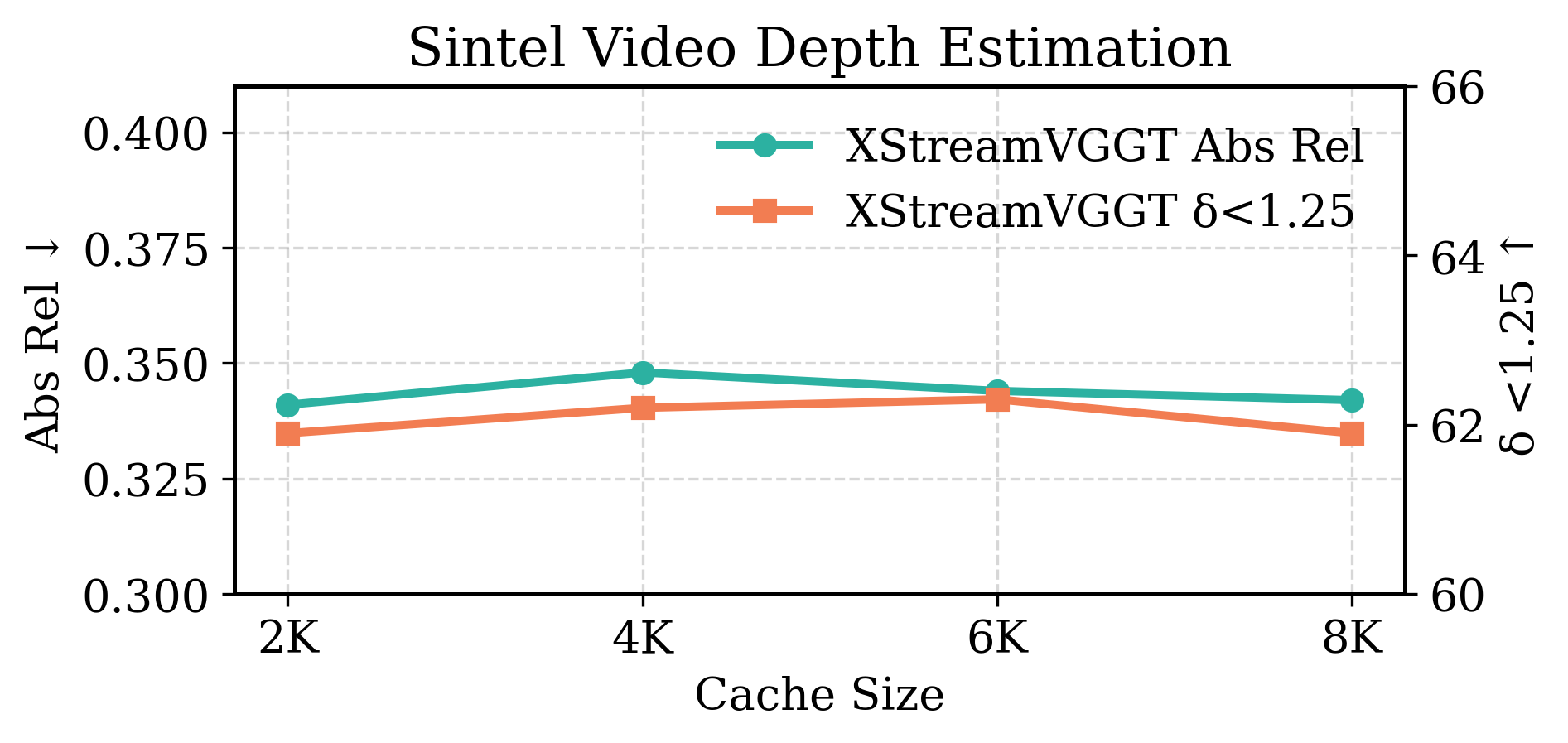}
\vspace{-5mm}
    \caption{Ablation study of cache length.}
    \label{fig:cache length}
    \end{subfigure}
    \begin{subfigure}{0.45\linewidth}
    \centering
    \includegraphics[width=\linewidth]{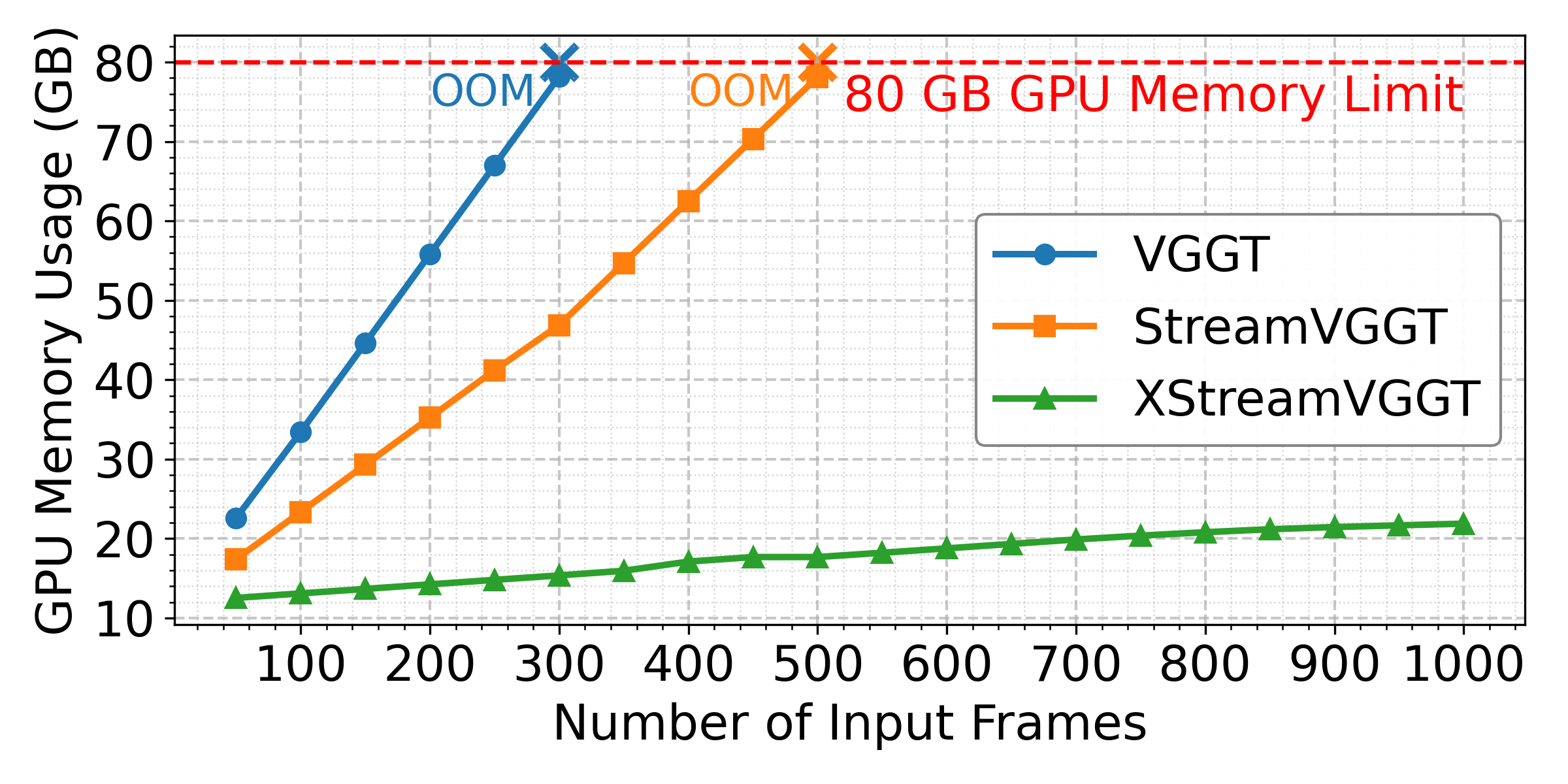}
\vspace{-5mm}
    \caption{Analysis of memory.}
    \label{fig:memory}
    \end{subfigure}
\caption{Ablation study of cache length and analysis of memory with increasing frame length.}
\vspace{-3mm}
\end{figure}

Figure~\ref{fig:cache length} presents an ablation study on cache length, with lengths varying across 2K, 4K, 6K, and 8K. 
As shown, performance changes minimally as cache length increases, suggesting significant redundancy among multi-frame patch tokens. 
A cache length of 2K provides strong performance while maintaining high efficiency, making it the optimal choice.
Additionally, we conduct an ablation study on both the pruning and quantization processes of XStreamVGGT to assess their individual effects. 
As indicated in Table~\ref{tab:ablation}, although pruning causes a slight performance decline, quantization introduces no additional degradation.

\subsection{Efficiency Analysis}
We evaluate the efficiency of XStreamVGGT using input sequences ranging from 50 to 1000 frames, measuring GPU memory consumption and inference speed. 
All experiments are conducted on a single 80GB A100 GPU. 
As shown in Figures~\ref{fig:fps} and~\ref{fig:memory}, both StreamVGGT and VGGT exhibit significant FPS degradation as the number of input frames increases and quickly encounter OOM errors. By contrast, XStreamVGGT consistently maintains substantially higher FPS without OOM, yielding significant efficiency gains with 4.42X lower memory usage and 5.48X faster inference compared with StreamVGGT.

\subsection{Qualitative Results}
\begin{figure}[t]
    \centering    

    \begin{subfigure}{0.45\columnwidth}
        \centering
    \includegraphics[width=\linewidth]{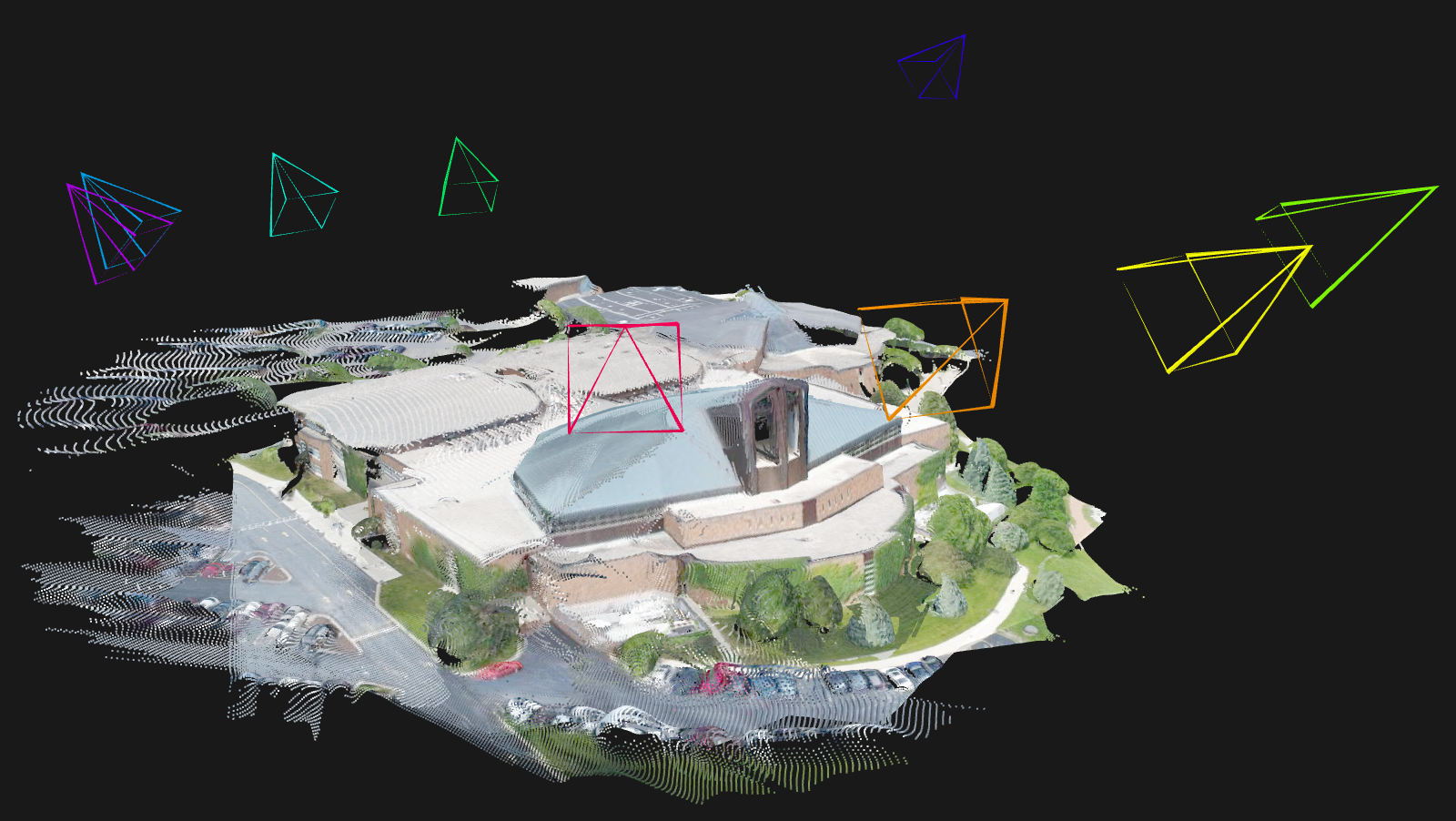}
    % %\vspace{-5mm}
    \caption{StreamVGGT.}
    \end{subfigure}
    \begin{subfigure}{0.45\columnwidth}
        \centering
    \includegraphics[width=\linewidth]{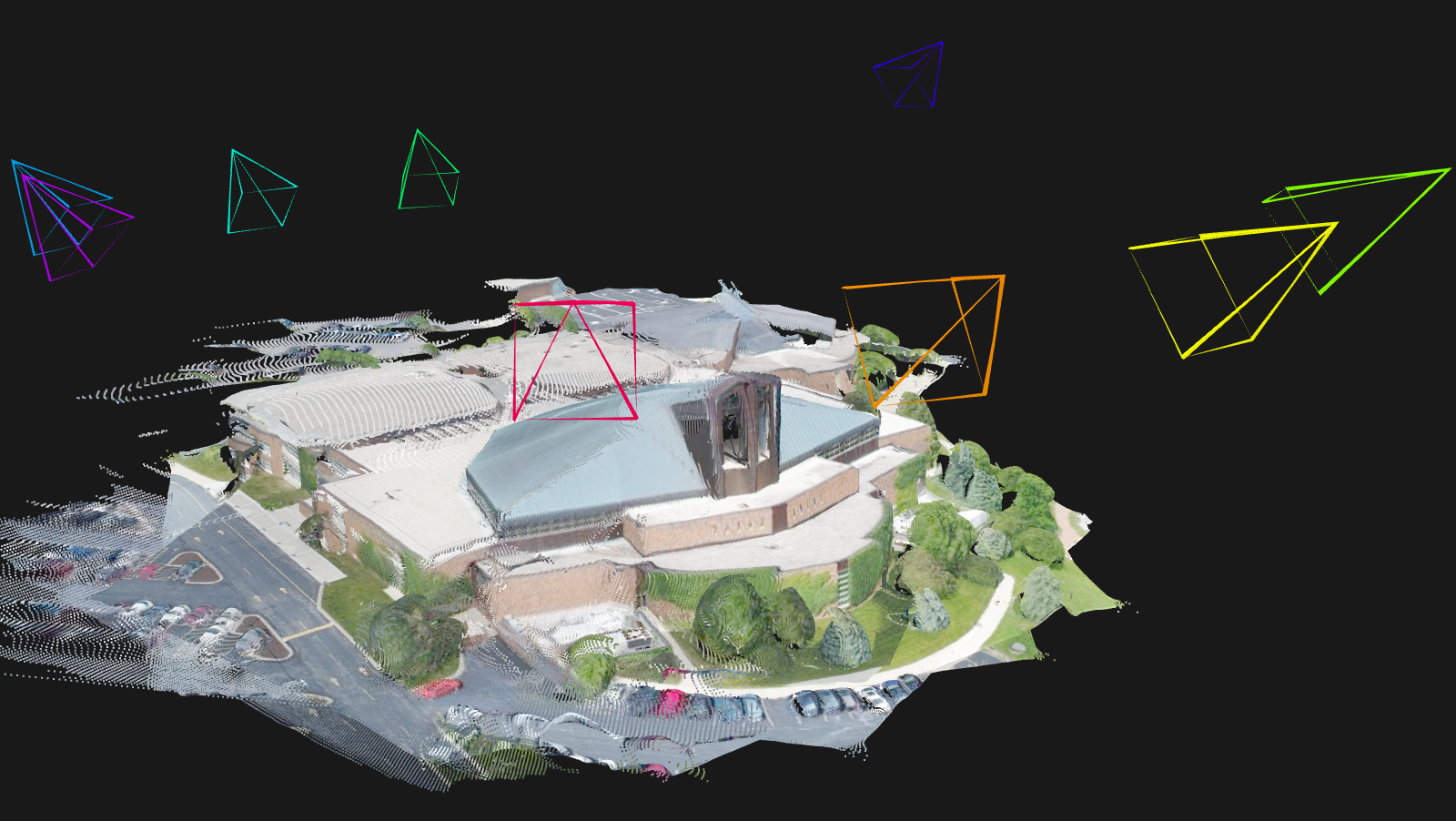}
    % %\vspace{-5mm}
    \caption{XStreamVGGT (ours).}
    \end{subfigure}

    \begin{subfigure}{0.45\columnwidth}
        \centering
    \includegraphics[width=\linewidth]{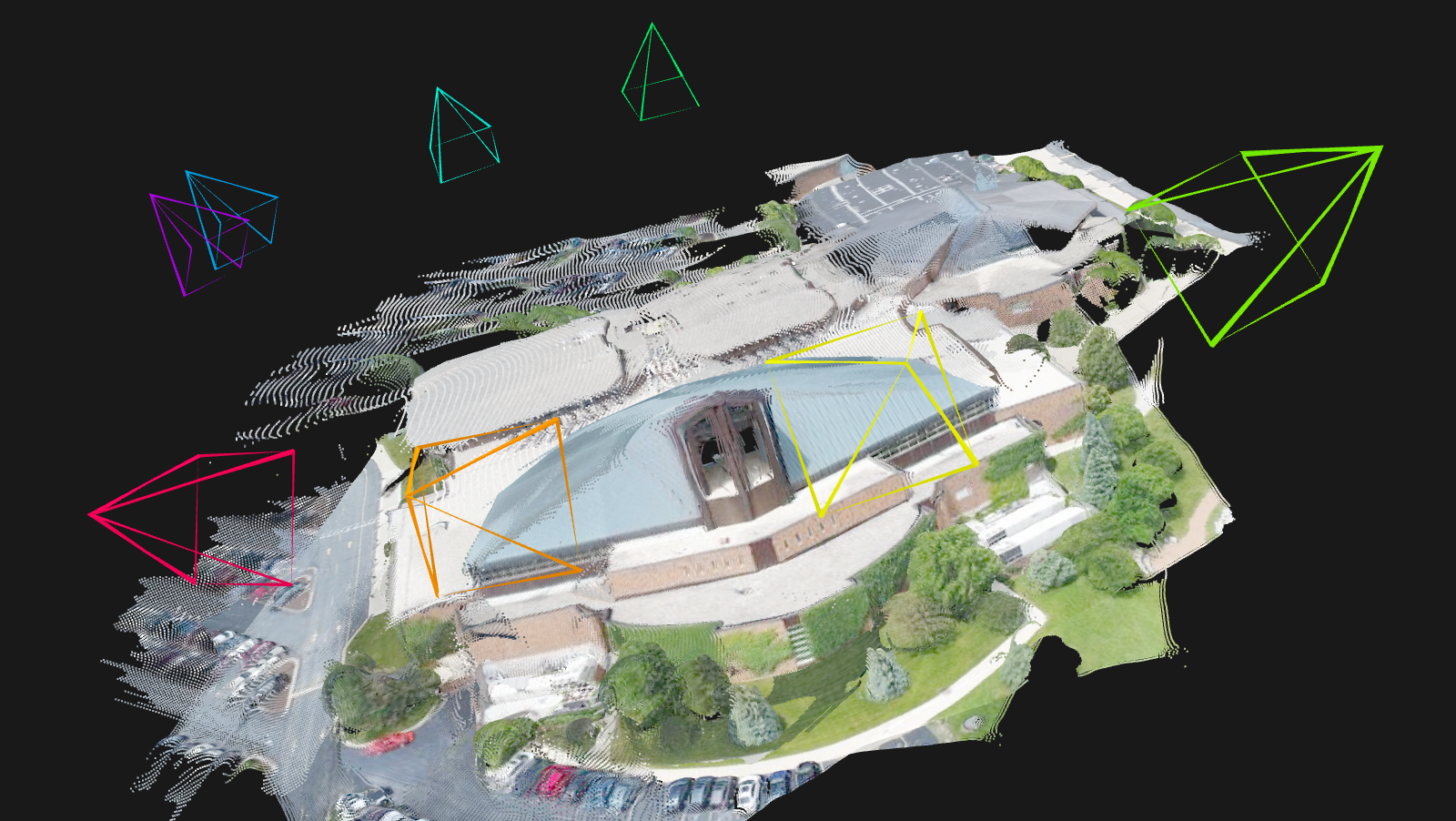}
    % %\vspace{-5mm}
    \caption{StreamVGGT.}
    \end{subfigure}
    \begin{subfigure}{0.45\columnwidth}
        \centering
    \includegraphics[width=\linewidth]{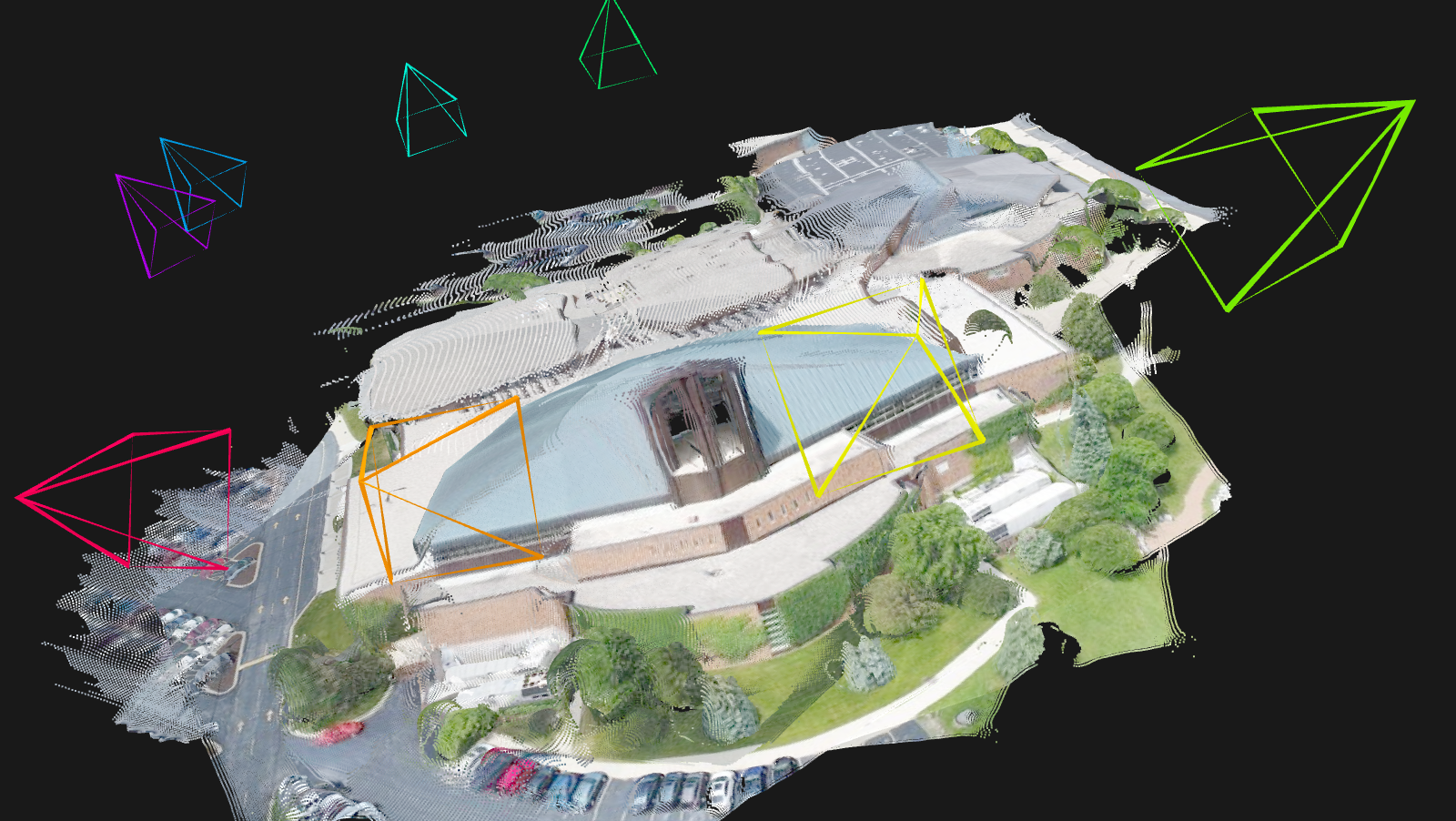}
    % %\vspace{-5mm}
    \caption{XStreamVGGT (ours).}
    \end{subfigure}

    % \begin{subfigure}{0.48\columnwidth}
    %     \centering
    % \includegraphics[width=\linewidth]{figure/recons/streamvggt_recons_3.png}
    % % %\vspace{-5mm}
    % \caption{StreamVGGT.}
    % \end{subfigure}
    % \begin{subfigure}{0.48\columnwidth}
    %     \centering
    % \includegraphics[width=\linewidth]{figure/recons/xstreamvggt_recons_3.png}
    % % %\vspace{-5mm}
    % \caption{XStreamVGGT (ours).}
    % \end{subfigure}

    \begin{subfigure}{0.45\columnwidth}
        \centering
    \includegraphics[width=\linewidth]{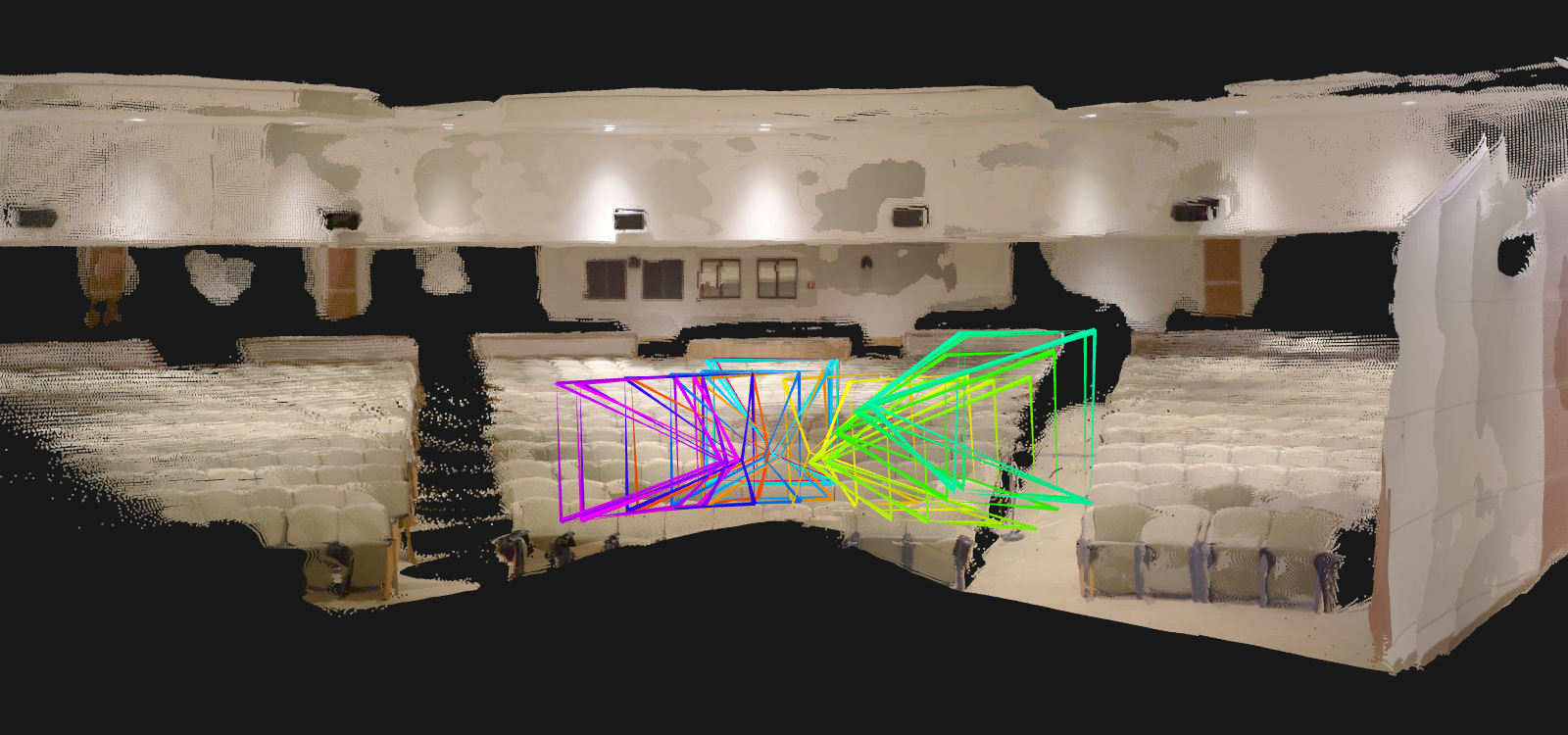}
    % %\vspace{-5mm}
    \caption{StreamVGGT.}
    \end{subfigure}
    \begin{subfigure}{0.45\columnwidth}
        \centering
    \includegraphics[width=\linewidth]{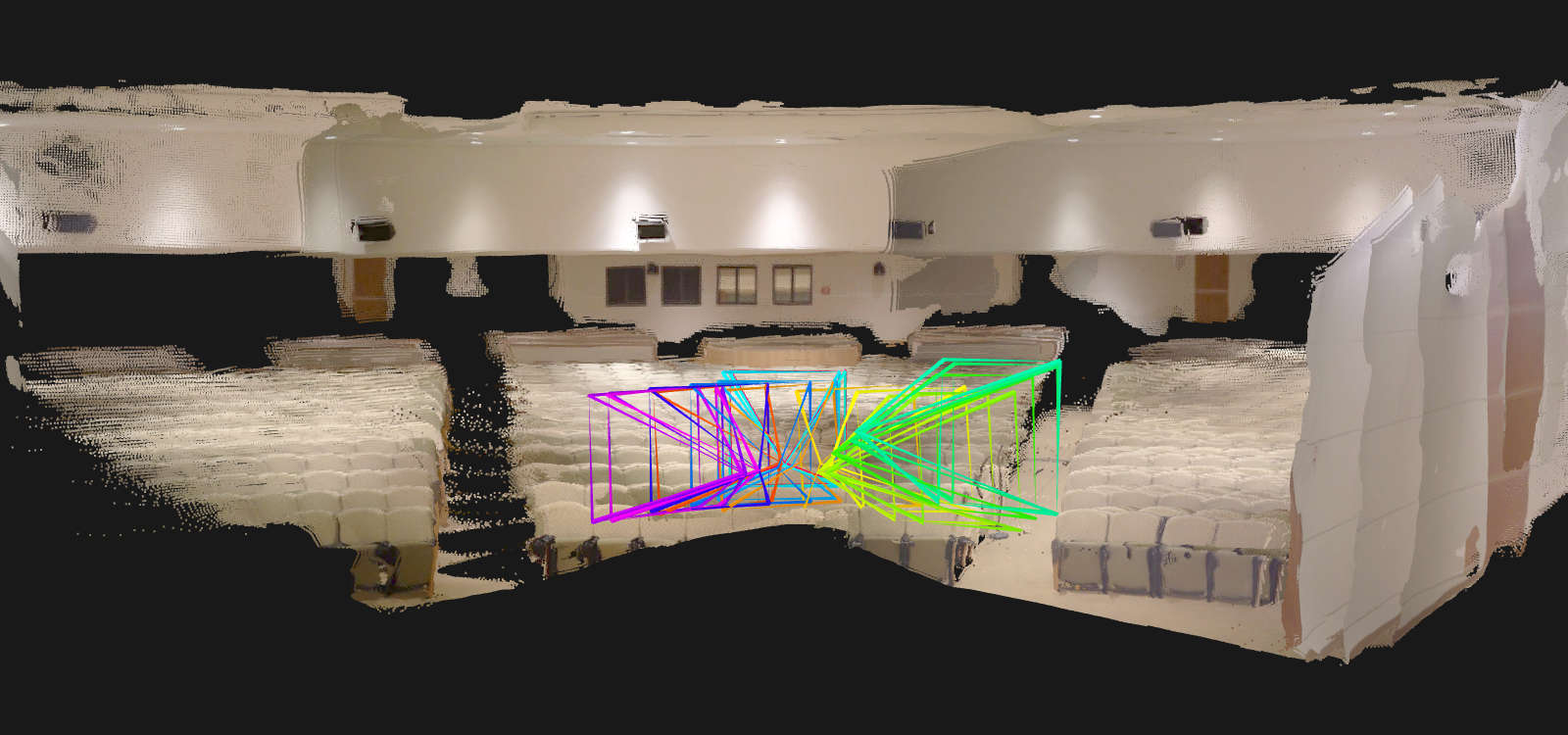}
    % %\vspace{-5mm}
    \caption{XStreamVGGT (ours).}
    \end{subfigure}

    \begin{subfigure}{0.45\columnwidth}
        \centering
    \includegraphics[width=\linewidth]{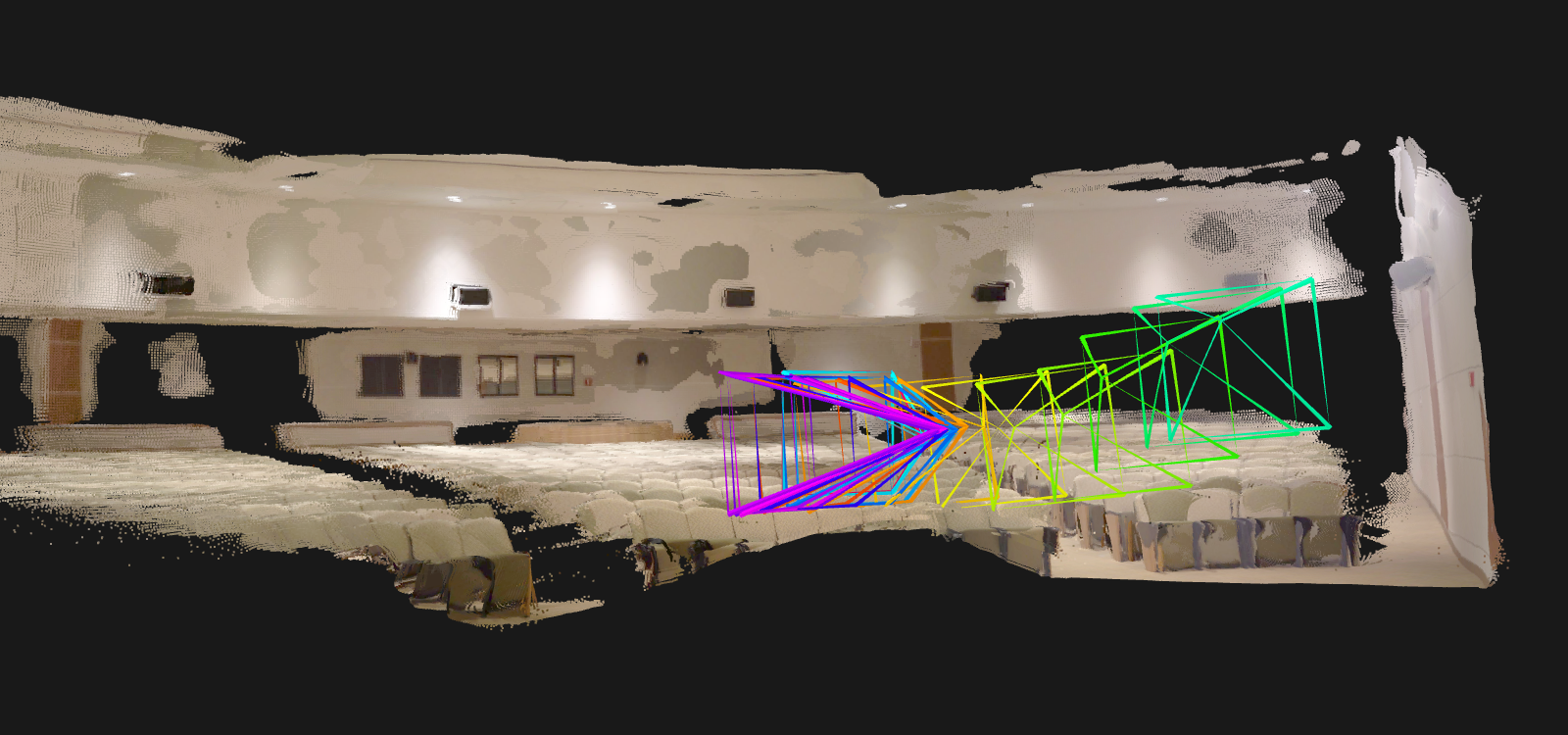}
    % %\vspace{-5mm}
    \caption{StreamVGGT.}
    \end{subfigure}
    \begin{subfigure}{0.45\columnwidth}
        \centering
    \includegraphics[width=\linewidth]{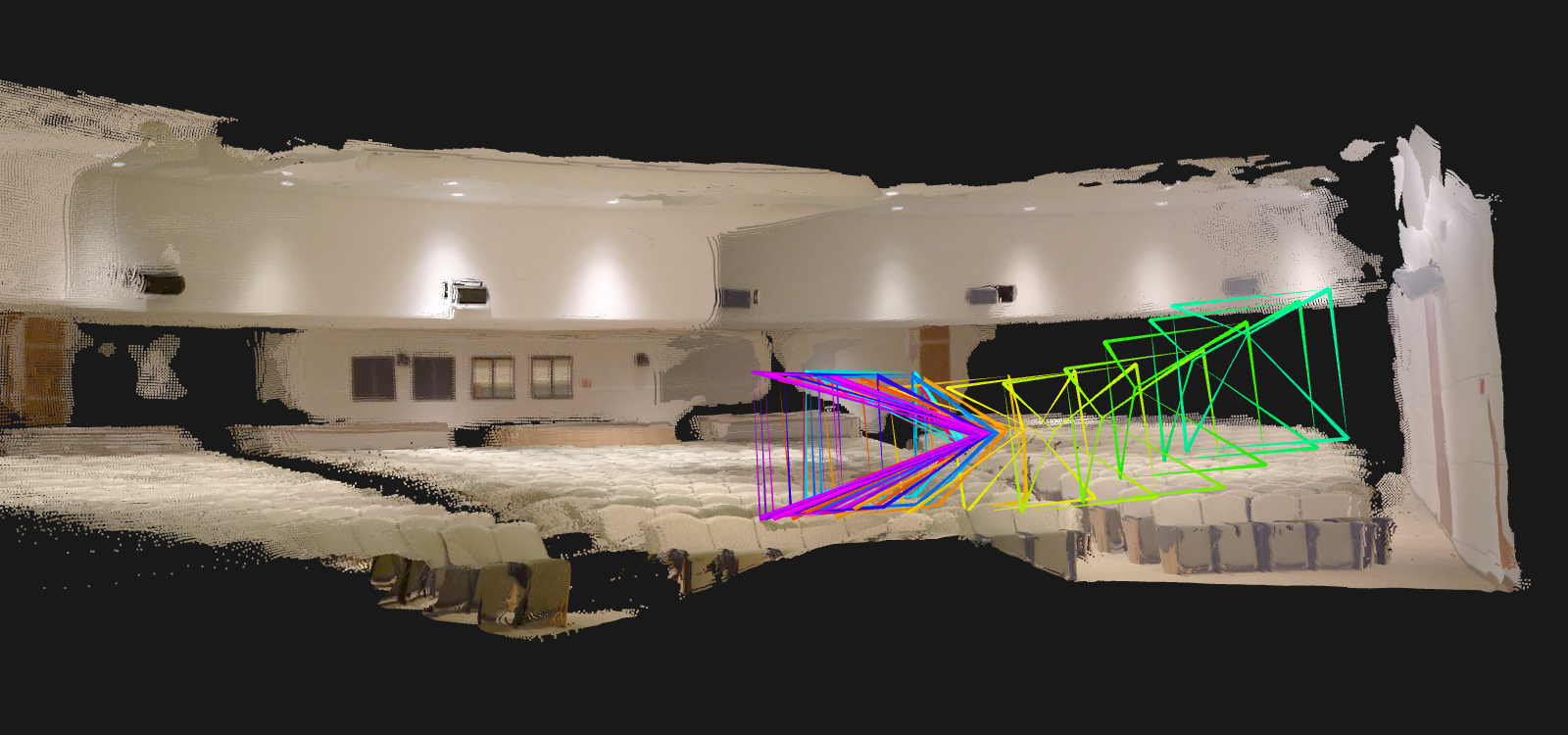}
    % %\vspace{-5mm}
    \caption{XStreamVGGT (ours).}
    \end{subfigure}

    \begin{subfigure}{0.45\columnwidth}
        \centering
    \includegraphics[width=\linewidth]{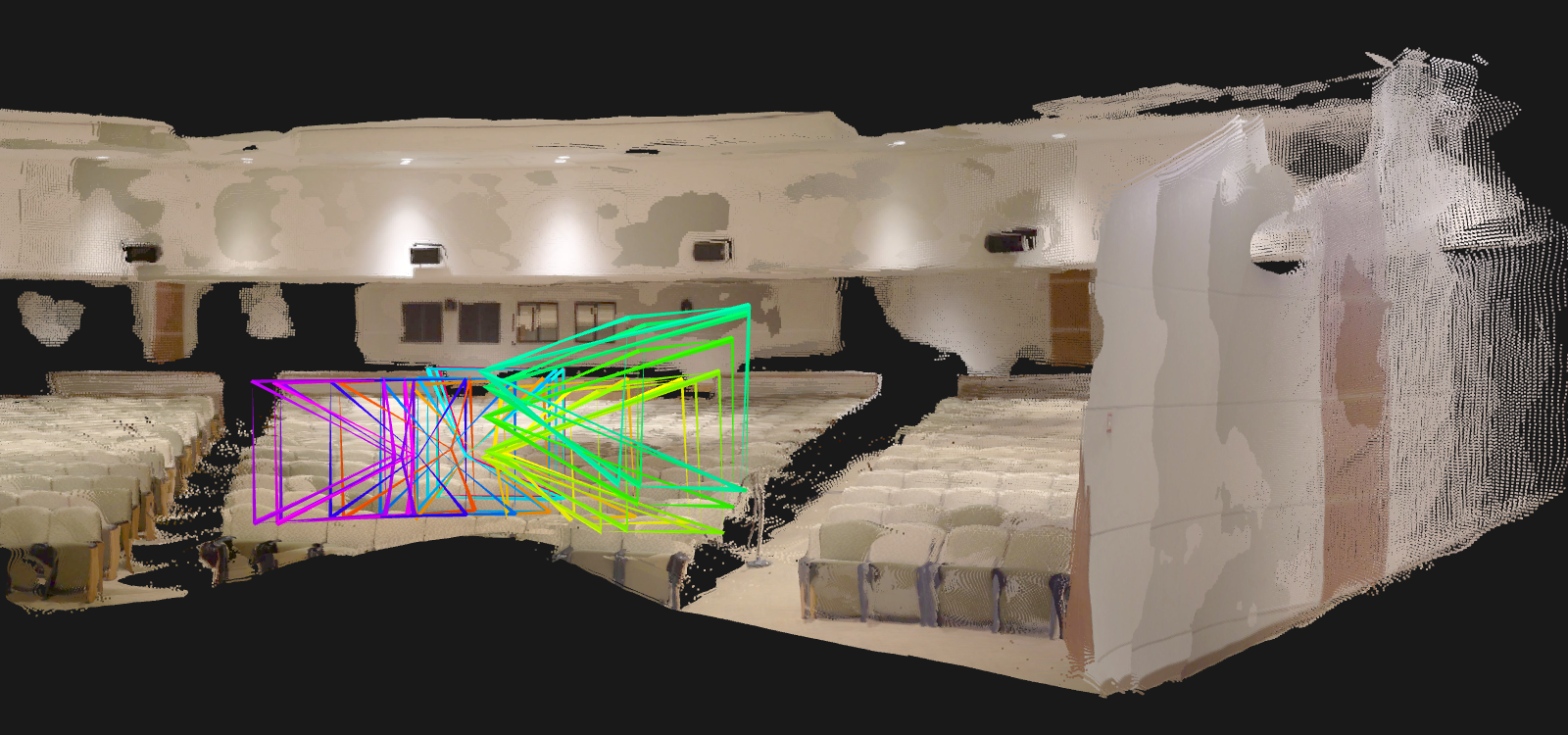}
    % %\vspace{-5mm}
    \caption{StreamVGGT.}
    \end{subfigure}
    \begin{subfigure}{0.45\columnwidth}
        \centering
    \includegraphics[width=\linewidth]{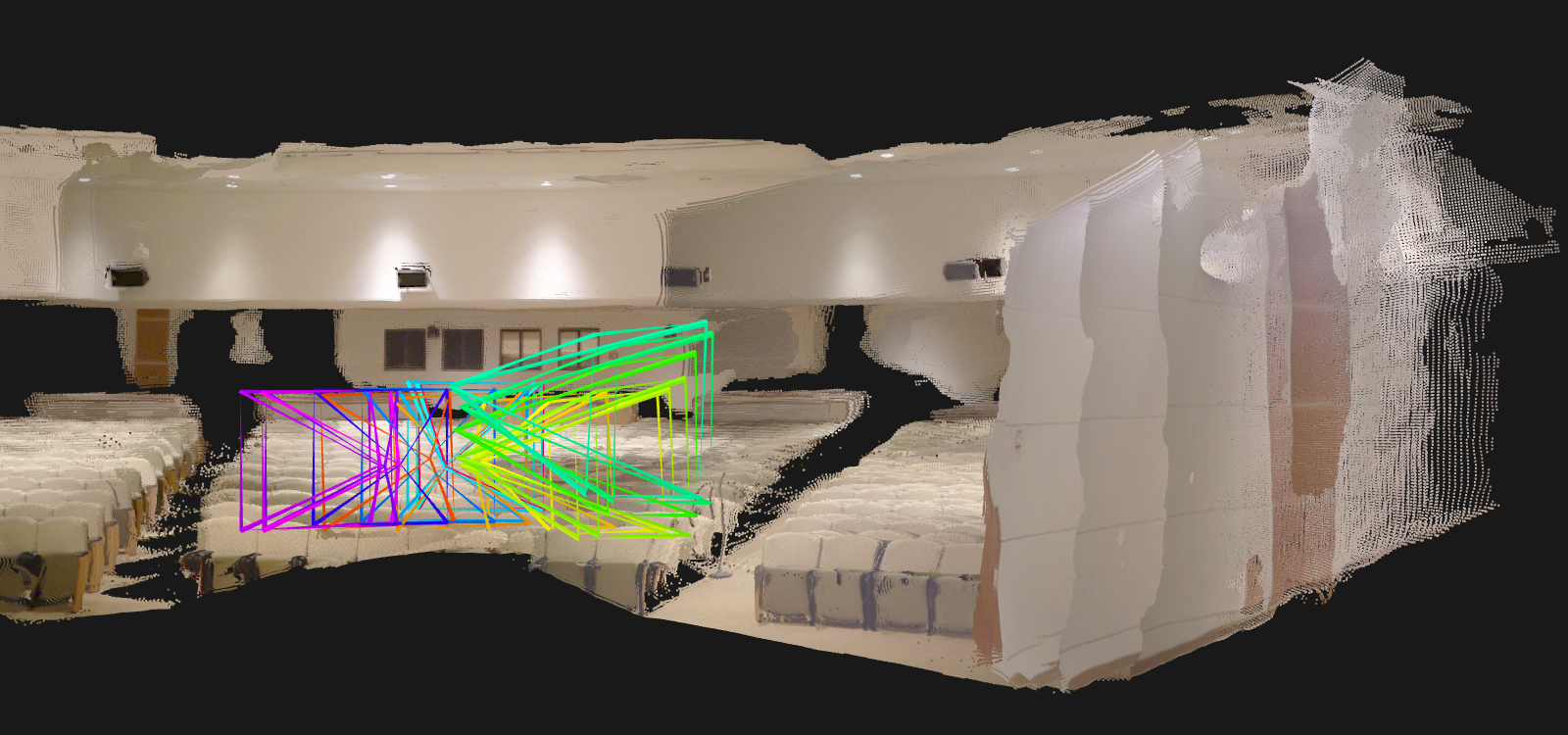}
    % %\vspace{-5mm}
    \caption{XStreamVGGT (ours).}
    \end{subfigure}

\caption{Qualitative reconstruction results comparing StreamVGGT and XStreamVGGT.}
    %\vspace{-2mm}
\label{fig:Qualitative results-1}
\end{figure}

\begin{figure}[t]
    \centering    

    \begin{subfigure}{0.48\columnwidth}
        \centering
    \includegraphics[width=\linewidth]{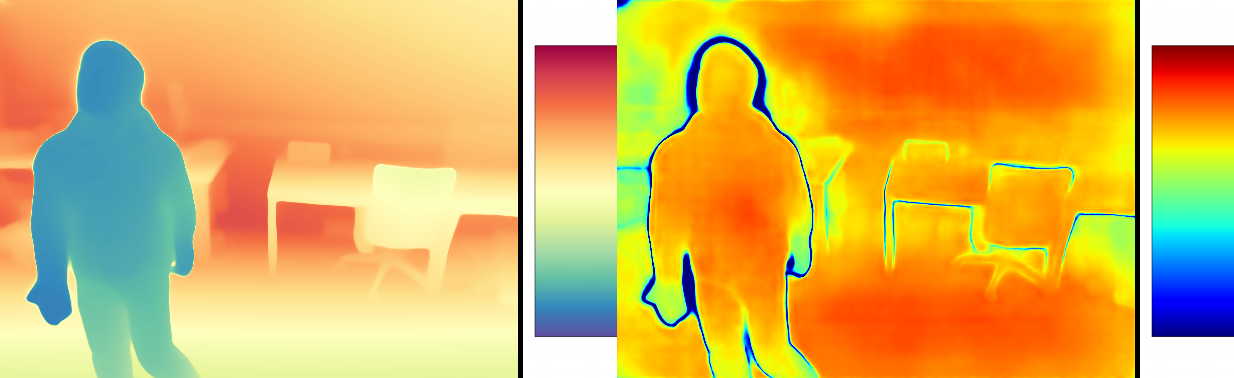}
    % %\vspace{-5mm}
    \caption{StreamVGGT.}
    \end{subfigure}
    \begin{subfigure}{0.48\columnwidth}
        \centering
    \includegraphics[width=\linewidth]{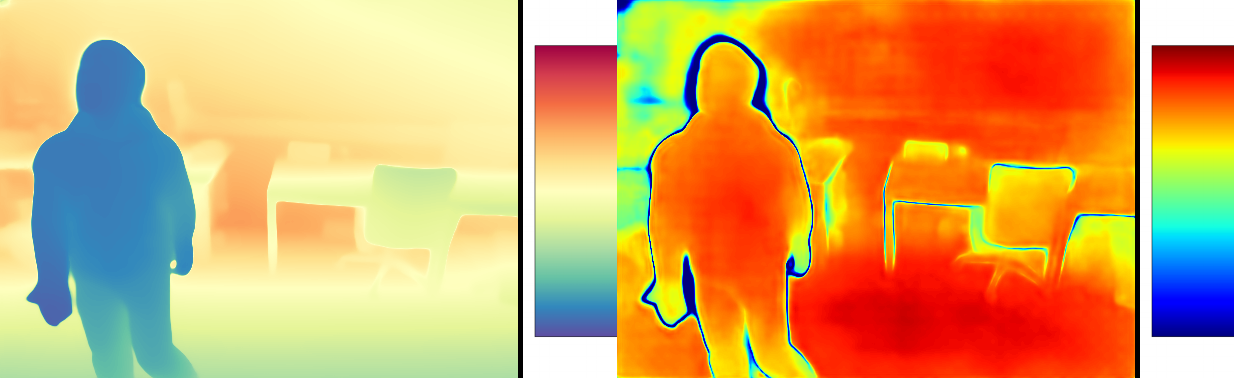}
    % %\vspace{-5mm}
    \caption{XStreamVGGT (ours).}
    \end{subfigure}

    \begin{subfigure}{0.48\columnwidth}
        \centering
    \includegraphics[width=\linewidth]{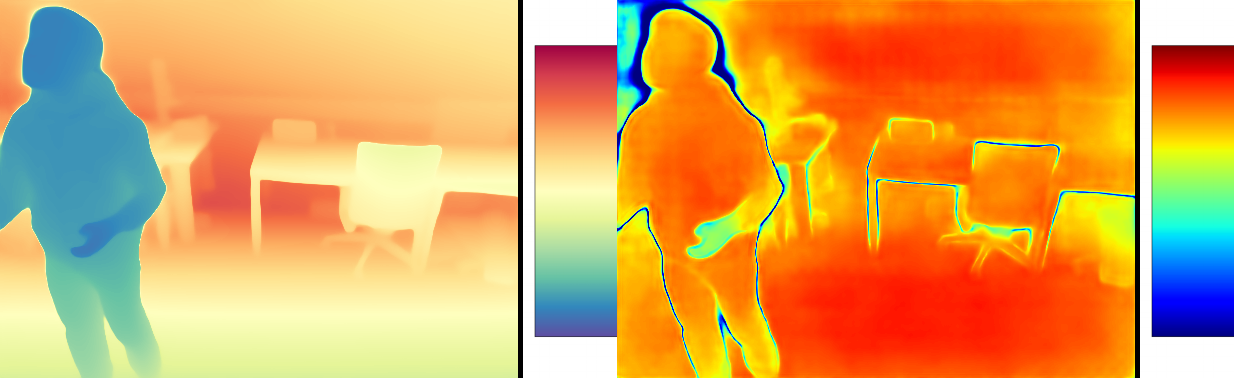}
    % %\vspace{-5mm}
    \caption{StreamVGGT.}
    \end{subfigure}
    \begin{subfigure}{0.48\columnwidth}
        \centering
    \includegraphics[width=\linewidth]{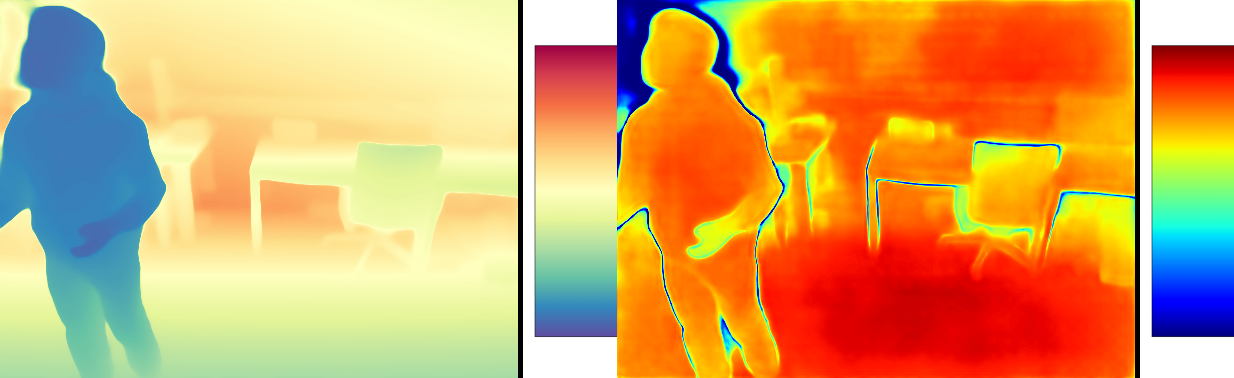}
    % %\vspace{-5mm}
    \caption{XStreamVGGT (ours).}
    \end{subfigure}

    \begin{subfigure}{0.48\columnwidth}
        \centering
    \includegraphics[width=\linewidth]{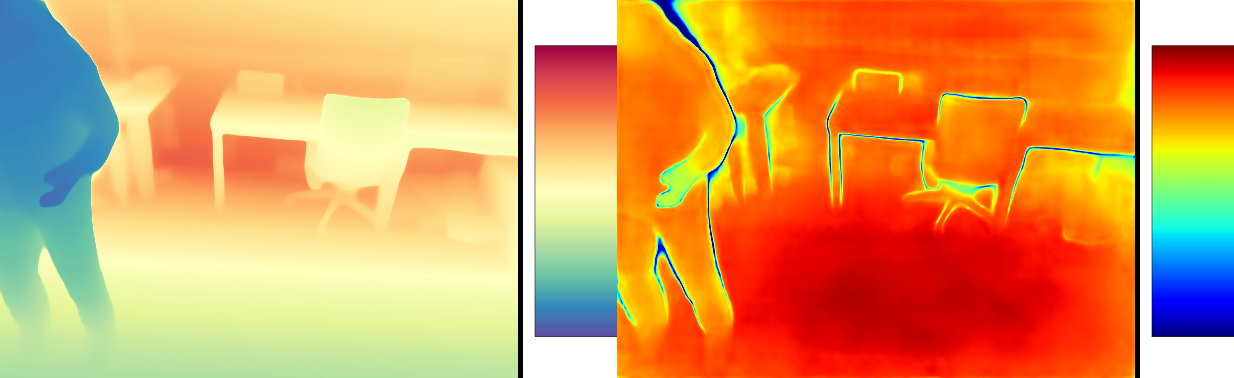}
    % %\vspace{-5mm}
    \caption{StreamVGGT.}
    \end{subfigure}
    \begin{subfigure}{0.48\columnwidth}
        \centering
    \includegraphics[width=\linewidth]{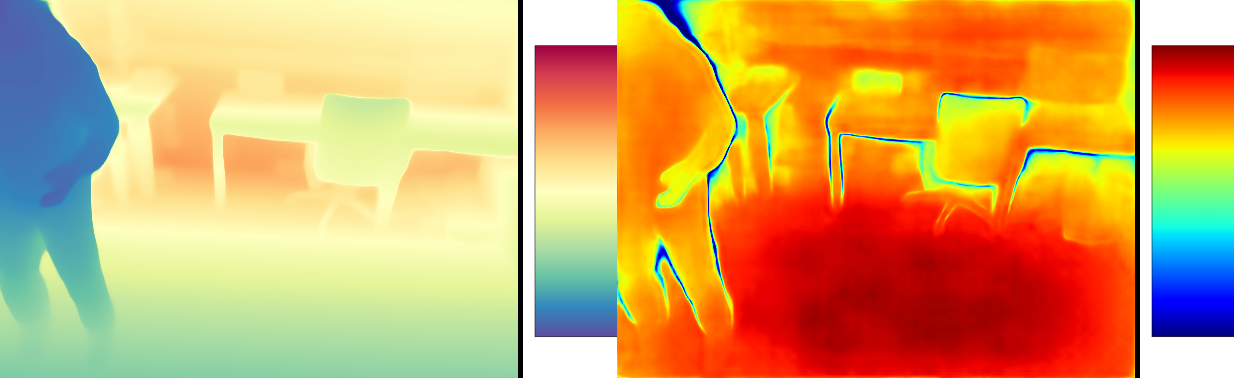}
    % %\vspace{-5mm}
    \caption{XStreamVGGT (ours).}
    \end{subfigure}

    \begin{subfigure}{0.48\columnwidth}
        \centering
    \includegraphics[width=\linewidth]{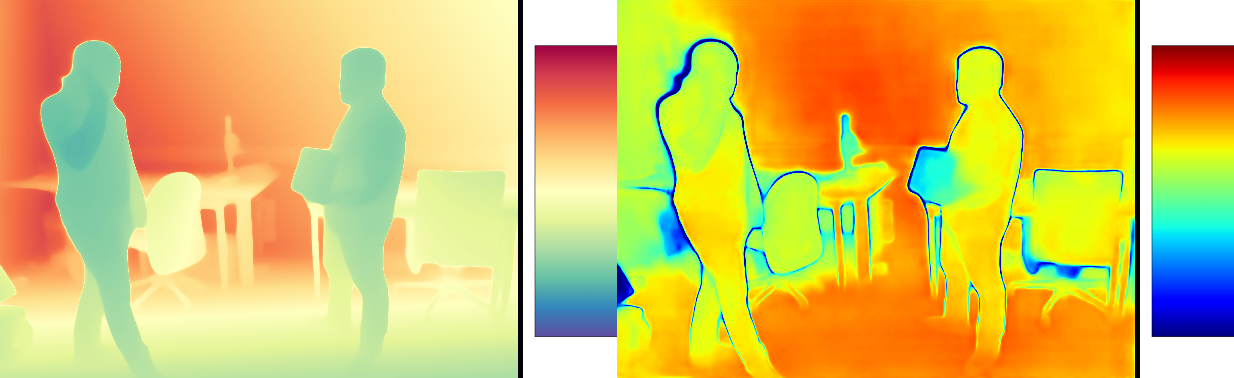}
    % %\vspace{-5mm}
    \caption{StreamVGGT.}
    \end{subfigure}
    \begin{subfigure}{0.48\columnwidth}
        \centering
    \includegraphics[width=\linewidth]{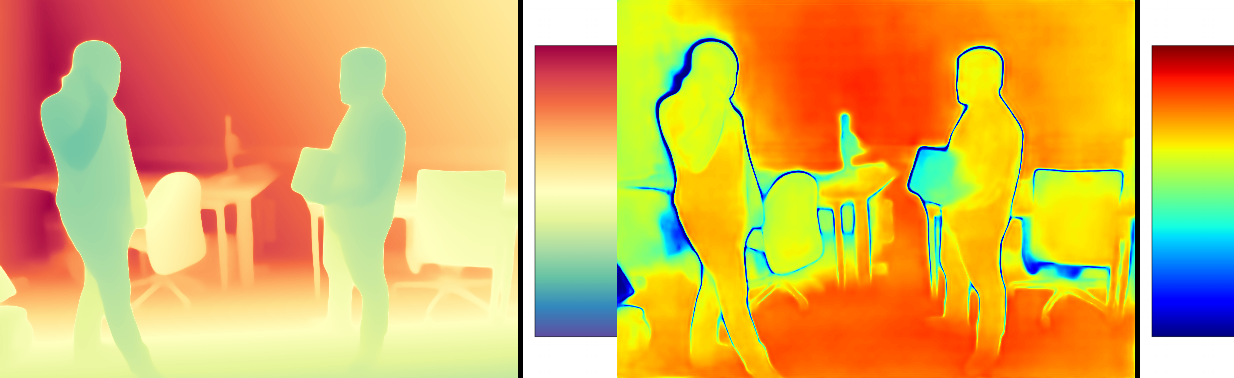}
    % %\vspace{-5mm}
    \caption{XStreamVGGT (ours).}
    \end{subfigure}

    \begin{subfigure}{0.48\columnwidth}
        \centering
    \includegraphics[width=\linewidth]{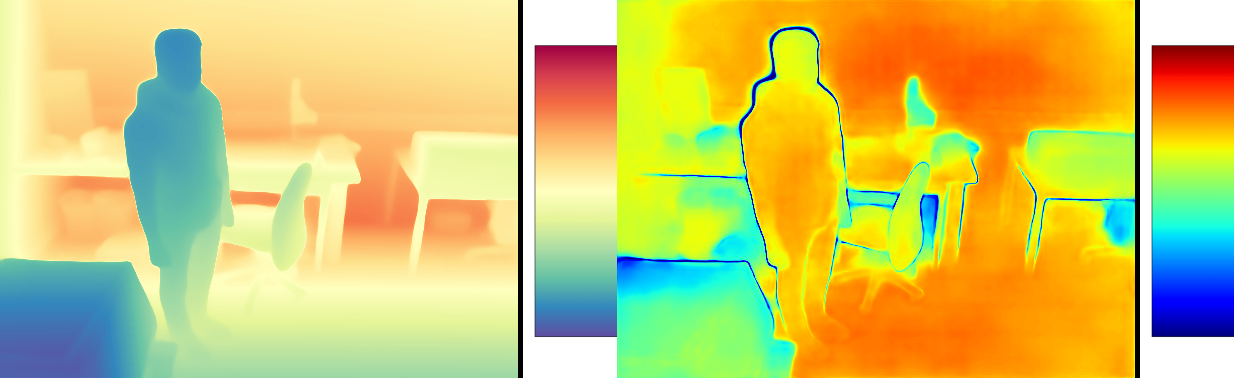}
    % %\vspace{-5mm}
    \caption{StreamVGGT.}
    \end{subfigure}
    \begin{subfigure}{0.48\columnwidth}
        \centering
    \includegraphics[width=\linewidth]{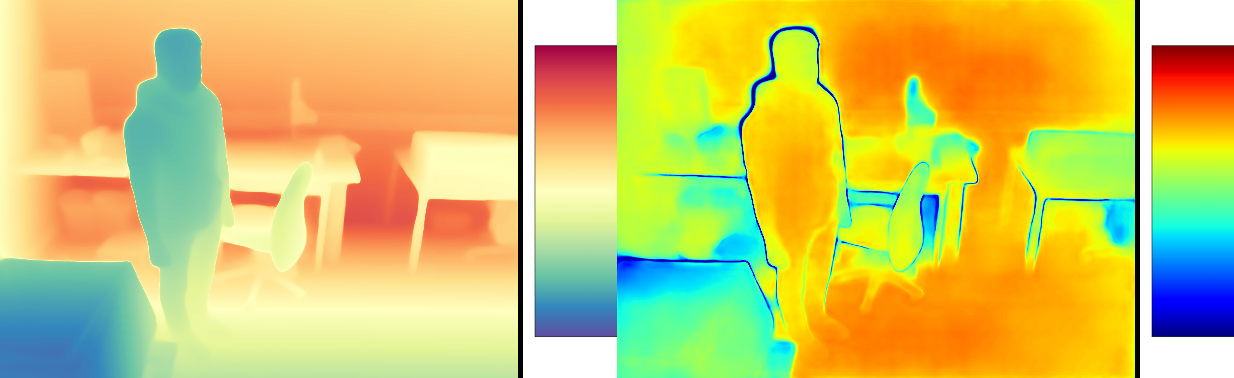}
    % %\vspace{-5mm}
    \caption{XStreamVGGT (ours).}
    \end{subfigure}

    \begin{subfigure}{0.48\columnwidth}
        \centering
    \includegraphics[width=\linewidth]{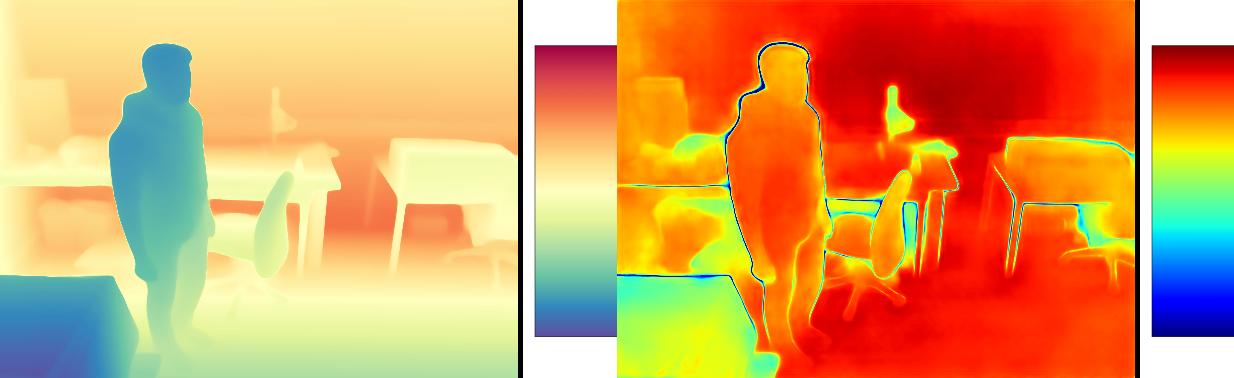}
    % %\vspace{-5mm}
    \caption{StreamVGGT.}
    \end{subfigure}
    \begin{subfigure}{0.48\columnwidth}
        \centering
    \includegraphics[width=\linewidth]{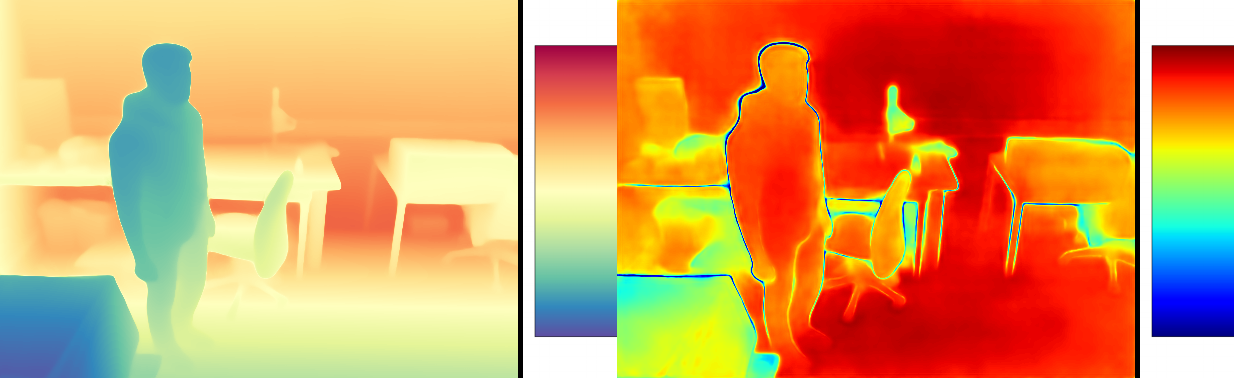}
    % %\vspace{-5mm}
    \caption{XStreamVGGT (ours).}
    \end{subfigure}

    \begin{subfigure}{0.48\columnwidth}
        \centering
    \includegraphics[width=\linewidth]{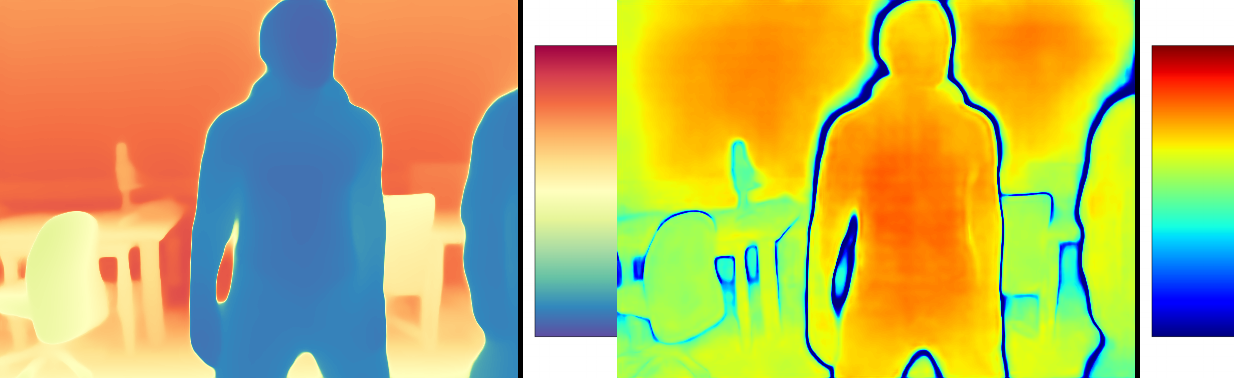}
    % %\vspace{-5mm}
    \caption{StreamVGGT.}
    \end{subfigure}
    \begin{subfigure}{0.48\columnwidth}
        \centering
    \includegraphics[width=\linewidth]{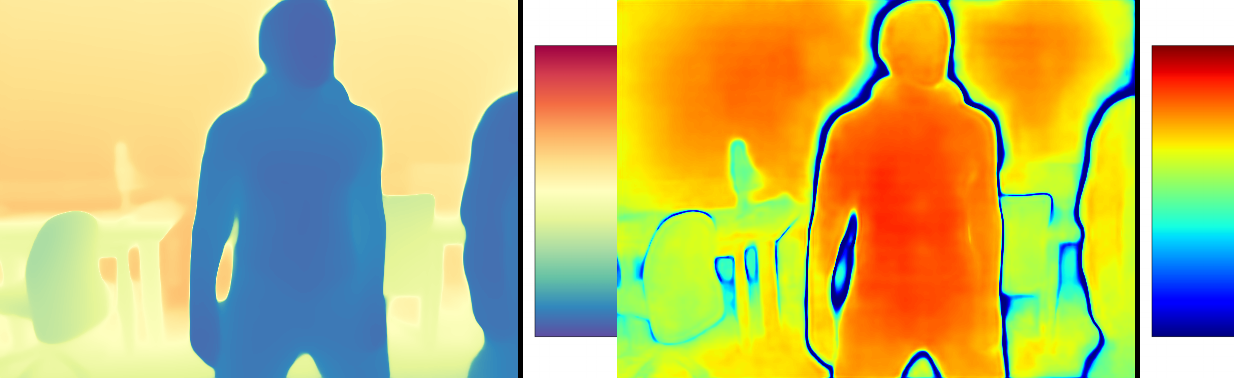}
    % %\vspace{-5mm}
    \caption{XStreamVGGT (ours).}
    \end{subfigure}

\caption{Qualitative depth estimation results comparing StreamVGGT and XStreamVGGT.}
    %\vspace{-2mm}
\label{fig:Qualitative results-2}
\end{figure}

Figures~\ref{fig:Qualitative results-1} and~\ref{fig:Qualitative results-2} present visual comparisons of 3D reconstruction and depth estimation results between StreamVGGT and XStreamVGGT. 
As shown, XStreamVGGT effectively preserves visual quality, producing results that closely match the original while delivering a notable improvement in computational efficiency.

\section{Conclusion}
We present XStreamVGGT, a tuning-free method for memory-efficient streaming inference of StreamVGGT.
By combining KV cache pruning with quantization, it bounds memory growth while preserving model fidelity. 
Extensive experiments demonstrate minimal performance loss alongside substantial reductions in memory footprint and inference latency, enabling scalable 3D streaming applications.
Future work will explore adaptive cache budgets that dynamically adjust based on scene complexity and motion characteristics.

\clearpage
% TBD

\clearpage
\section{Acknowledgments}
This work was supported in part by the Research Grants Council of the Hong Kong Special Administrative Region Government through the TRS project T45-701/22-R and GRF project 17203224, and the TCL Corporate Research (Hong Kong) Co., Limited.

% \bibliography{neurips_2025}

\bibliographystyle{plain}

\end{document}